\pdfoutput=1

\RequirePackage[2020-02-02]{latexrelease}
\documentclass{clv3}

\usepackage{hyperref}
\usepackage{xcolor}
\definecolor{darkblue}{rgb}{0, 0, 0.5}
\hypersetup{colorlinks=true,citecolor=darkblue, linkcolor=darkblue, urlcolor=darkblue}

\usepackage{algorithmic}

\usepackage[utf8]{inputenc}
\usepackage[english]{babel}                     

\usepackage{natbib}
\usepackage[nolist,nohyperlinks]{acronym}       
\usepackage{footnote}                           
\usepackage{url}                                

\usepackage{algorithm} 
\usepackage{booktabs}
\usepackage{multirow, tabularx}                 
\usepackage{tabulary}                           

\usepackage{amsopn}

\usepackage{array}

\usepackage{adjustbox}                          
\usepackage{pdflscape}                          

\newcommand{\STAB}[1]{\begin{tabular}{@{}c@{}}#1\end{tabular}}

\graphicspath{{./img/}}

\begin{document}

\issue{xx}{yy}{2022}

\dochead{Long Paper}

\runningtitle{Active Learning with Expert Advice for Real World MT}

\runningauthor{Mendonça et al}

\title{{\em Onception}: Active Learning with Expert Advice for Real World Machine Translation}

\author{Vânia Mendonça\thanks{Corresponding author. E-mail: vania.mendonca@tecnico.ulisboa.pt}}
\affil{INESC-ID, Instituto Superior Técnico}

\author{Ricardo Rei}
\affil{INESC-ID, Instituto Superior Técnico, Unbabel AI}

\author{Luísa Coheur}
\affil{INESC-ID, Instituto Superior Técnico}

\author{Alberto Sardinha}
\affil{INESC-ID, Instituto Superior Técnico}

\maketitle

\begin{abstract}

Active learning can play an important role in low-resource settings (i.e., where annotated data is scarce), by selecting which instances may be more worthy to annotate. Most active learning approaches for Machine Translation assume the existence of a pool of sentences in a source language, and rely on human annotators to provide translations or post-edits, which can still be costly. 
In this article, we assume a real world human-in-the-loop scenario in which: (1) the source sentences may not be readily available, but instead arrive in a stream; (2) the automatic translations receive feedback in the form of a rating, instead of a correct/edited translation, since the human-in-the-loop might be a user looking for a translation, but not be able to provide one. To tackle the challenge of deciding whether each incoming pair source-translations is worthy to query for human feedback, we resort to a number of stream-based active learning query strategies. Moreover, since we not know in advance which query strategy will be the most adequate for a certain language pair and set of Machine Translation models, we propose to dynamically combine multiple strategies using prediction with expert advice. 
Our experiments show that using active learning allows to converge to the best Machine Translation systems with fewer human interactions. Furthermore, combining multiple strategies using prediction with expert advice often outperforms several individual active learning strategies with even fewer interactions.
\end{abstract}



\begin{acronym}[EEGPET]

    \acrodefplural{RQ}[RQ]{research questions}
    
    \acro{CRF}      {Conditional Random Fields}
    \acro{EGAL}     {Exploration-Guided Active Learning}
    \acro{EWAF}     {Exponentially Weighted Average Forecaster}
    \acro{EXP4}     {Exponential-weighting for Exploration and Exploitation with Experts}
    \acro{EXP3}     {Exponential-weighting for Exploration and Exploitation}
    \acro{$F_1$}    {F-Score}
    \acro{FN}       {false negatives}
    \acro{FP}       {false positives}
    \acro{LDC}      {Linguistic Data Consortium}
    \acro{HMM}      {Hidden Markov Model}
    \acro{IDen}     {Information Density}
    \acro{IL}       {Imitation Learning}
    \acro{IMT}      {Interactive Machine Translation}
    \acro{LTAL}     {Learning To Active Learn}
    \acro{LC}       {Least Confidence}
    \acro{MDP}      {Markov Decision Problem}
    \acro{MT}       {Machine Translation}
    \acro{MQM}      {Multidimensional Quality Metric}
    \acro{NER}      {Named Entity Recognition}
    \acro{NLP}      {Natural Language Processing}
    \acro{NLU}      {Natural Language Understanding}
    \acro{POMDP}    {Partially Observable Markov Decision Problem}
    \acro{PoS}      {Part-of-Speech}
    \acro{QbC}      {Query-by-Committee}
    \acro{QE}       {Quality Estimation}
    \acro{RL}       {Reinforcement Learning}
    \acro{RQ}       {research question}
    \acro{StDev}    {Standard Deviation}
    \acro{TN}       {true negatives}
    \acro{TP}       {true positives}
    \acro{US}       {Uncertainty Sampling}
    \acro{UCB}      {Upper Confidence Bound}
    \acro{WM}       {Weighted Majority}
    \acro{WMT}      {Conference on Machine Translation}

\end{acronym}

\section{Introduction}
\label{sec:intro}

The state of the art on most \acs{NLP} tasks has been dominated by supervised neural approaches, and \ac{MT} is no exception to this trend \citep{barrault-etal-2020-findings}. The impressive results obtained by neural models became possible due to the growing amount of annotated data available; however, this growth is not observed for most languages, since annotation (or, in the case of \ac{MT}, translation) is a time-consuming and expensive process. This motivates the use of {\em low-resource} learning methods (i.e., methods that can make the most of minimal annotation effort). 
Active learning is one of such methods, as it allows to choose which instances should be annotated, based on some criterion that measures their informativeness \citep{Cohn1994,Settles2010}. Active learning has been extensively applied to \ac{MT} \citep[e.g.][]{Haffari2009,Ambati2011,Gonzalez-Rubio2014,Zeng2019}, mostly in a pool-based setting (i.e., when a pool of source sentences is available and one has to choose which sentences should be translated by a human annotator). 

Existing active learning approaches for \ac{MT} assume a scenario where a human annotator either provides the translations for sentences in the source language or post-edits the translations outputted by an \ac{MT} model (in the case of Interactive \ac{MT}). To the best of our knowledge, no works have explored a real world scenario in which users of \ac{MT} systems (e.g., Web \ac{MT} services) could themselves be a source of feedback. In this case, the human would be looking for a translation, rather than providing it, but they might still be able to provide feedback (e.g., in the form of a rating) on whether the automatic translation makes sense in the target language. Such ratings could be useful for fine-tuning an \ac{MT} model (or an ensemble of models) and require considerably less effort than providing a translation from scratch or post-editing a translation. 

To fill this gap, we build on our previous work \citep{Mendonca2021}, in which we leveraged on human ratings to learn the weights of an ensemble of arbitrary \ac{MT} models in an online fashion, in order to dynamically improve its performance for the language pairs in hand. We extend this proposal by incorporating model-independent active learning strategies that decide, for each pair source-translations, whether it is worthy to query the user for a rating, thus reducing the effort needed to improve the performance of the \ac{MT} ensemble. 

Moreover, since we do not know in advance which strategies will be the most adequate for a certain language pair and ensemble of models, we dynamically combine multiple active learning strategies under the framework of prediction with expert advice, following another previous work of ours that focused on sequence labeling tasks \citep{Mendonca2020}. 
Thus, we introduce a second layer of online learning\footnote{Thus {\em Onception}, in the sense of ``dream within a dream'' used in the movie {\em Inception}.} that learns the strategies' weights based on the performance of the online \ac{MT} ensemble at each user interaction. 

We thus address the following research questions:
\begin{description}
    \item[RQ1] Can an active learning approach find the best systems with fewer human interactions?
    \item[RQ2] Does it pay off to combine multiple query strategies using expert advice (rather than using a single strategy)?
\end{description}

Our contribution is four-fold: 

\begin{enumerate}
\item A model-independent active learning solution that makes the most of an ensemble of pre-trained models by taking advantage of reduced human effort to dynamically adapt the ensemble to any language pair in hand\footnote{The code for our experiments can be found in \url{https://github.com/vania-mendonca/Onception}};
\item A set of experiments in a much lower-resource setting than most existing related literature, as we use datasets containing only as many as 1000 to 2000 source sentences;
\item A set of stream-based active learning strategies that do not require any additional pre-training data (unlike recent active learning approaches to \ac{MT} \citep{Liu2018,Finkelstein2020}); 
\item A novel application of prediction with expert advice and active learning to \ac{MT} in which we dynamically combine multiple stream-based strategies (instead of committing to a single strategy a priori).
\end{enumerate}

Our experiments on \acs{WMT}'19 datasets \citep{Barrault2019} show that using active learning indeed allows to converge to the top rated \ac{MT} systems for each language pair, while sparing human interactions. Moreover, since the best query strategy varies across language pairs and online learning settings, combining multiple active learning strategies using prediction with expert advice is generally a safer option than committing to a single strategy, and often outperforms several individual strategies.

The remainder of this article is structured as follows: in Section~\ref{sec:bg}, we provide some background on the frameworks of active learning and prediction with expert advice; in Section~\ref{sec:rw}, we present related work regarding the use of online and active learning in \ac{MT}; in Section~\ref{sec:problem}, we provide the starting point of our work and define the problem we are tackling; in Section~\ref{sec:ALxOLxMT}, we propose a set of stream-based active learning strategies to reduce human intervention while improving an \ac{MT} ensemble; in Section~\ref{sec:OLxALxOLxMT}, we propose the combination of multiple active learning strategies under the framework of prediction with expert advice; in Section~\ref{sec:setup}, we present the details of our experiments, whose results are reported and discussed in Section~\ref{sec:results}; finally, in Section~\ref{sec:cfw}, we wrap-up this article and present future work directions.

\section{Background}
\label{sec:bg}

In this section, we provide the necessary background on the learning frameworks applied in this work: active learning (Section~\ref{sec:AL}) and  prediction with expert advice (Section~\ref{sec:experts}). 


\subsection{Active Learning}
\label{sec:AL}

Active learning is a learning framework that aims at minimizing the amount of annotated data needed to train a supervised model, by choosing which instances should be annotated, based on some criterion that measures their informativeness ({\bf query strategy}) \citep{Cohn1994,Settles2010}. 

Active learning is most commonly used in a {\bf pool-based} setting (see Algorithm~\ref{alg:ALpool}): a model is trained on an initially small labeled set $\mathcal{L}$ and further retrained as $\mathcal{L}$ is iteratively augmented with an instance (or a batch of instances) selected from a pool of unlabeled instances $\mathcal{U}$. Each instance (or batch) $u_t$ is selected according to a query strategy (line~\ref{l:qspool}) and sent to be annotated by a human (line~\ref{l:humanpool}). The now annotated instance(s) $u_t^*$ are removed from $\mathcal{U}$ and added to $\mathcal{L}$, and this process repeats itself until a budget of $T$ instances is exhausted. 

\begin{algorithm}
\caption{Pool-based active learning}
\label{alg:ALpool}
\begin{algorithmic}[1]
    \REQUIRE model, labeled set $\mathcal{L}$, unlabeled set $\mathcal{U}$, budget $T$, batch size $B$
    
    \FOR{$t \gets 1$ to $T$}

        \STATE $model.train(\mathcal{L})$
        \STATE $\hat{Y} \gets model.predict(\mathcal{U})$

        \STATE $u_t \gets selectInstances(\mathcal{U}, \hat{Y}, \mathcal{L}, B)$  \label{l:qspool}

        \STATE $u_t^* \gets askAnnotation(u_t) $ \label{l:humanpool}

        \STATE $\mathcal{L} \gets \mathcal{L} \cup u_t^*$

        \STATE $\mathcal{U} \gets \mathcal{U} - u_t$

    \ENDFOR

\end{algorithmic}
\end{algorithm}

However, there might be situations in which one does not have access in advance to a pool of unlabeled instances, but rather accesses them progressively (consider, for example, an interactive system). In such situations, a {\bf stream-based} setting can be followed instead (see Algorithm~\ref{alg:ALstream}). In this setting, for each unlabeled instance (or batch of instances) $u_t$ that occurs in the stream, the query strategy has to decide whether it is worthy to ask the human for its annotation (line~\ref{l:qsstream}). 

\begin{algorithm}
\caption{Stream-based active learning}
\label{alg:ALstream}
\begin{algorithmic}[1]
    \REQUIRE model, labeled set $\mathcal{L}$, stream of unlabeled instances $\mathcal{S}$

    \FORALL{$u_t \in \mathcal{S}$}

        \STATE $\hat{y}_t \gets model.predict(u_t)$

        \IF{$selectInstance?(u_t, \hat{y}_t)$} \label{l:qsstream}

            \STATE $u_t^* \gets askAnnotation(u_t)$
            \STATE $\mathcal{L} \gets \mathcal{L} \cup u_t^*$
            \STATE $model.train(\mathcal{L})$

        \ENDIF
    \ENDFOR

\end{algorithmic}
\end{algorithm}

Active learning query strategies can be based on information related to the learning models, on the characteristics of the data, or a mix of both. {\bf Model-based} strategies rely on criteria such as the model's confidence on its prediction (e.g., Uncertainty Sampling \citep{Lewis1994}), the disagreement among the predictions of different models (e.g., Query-by-Committee \citep{Seung1992}), or the change expected in the model after being retrained with the selected instances (e.g., Fisher Information \citep{Settles2008}). 
{\bf Data-based} strategies, on the other hand, focus on the representativeness of the instances -- either how similar an instance is to the unlabeled set $\mathcal{U}$ (Density), or how much it differs from the labeled set $\mathcal{L}$ (Diversity) \citep{Fujii1998}, with some strategies combining both criteria (e.g., Exploration-Guided Active Learning \citep{Hu2010}). 
Finally, a popular strategy that combines both the model uncertainty about its predictions and the instances' Density with respect to $\mathcal{U}$ is \acl{IDen} \citep{Settles2008}. In the case of a stream-based setting, in which the decision on whether to select an instance cannot depend on the remaining unlabeled instances, query strategies may rely on a threshold to make its decision instead. 

More recently, several works have proposed to learn the query strategy from data (known as {\bf \acl{LTAL}}), instead of using the previously mentioned heuristics (or in combination with them). These works framed the problem of selecting an instance to be annotated as a regression problem \citep{Konyushkova2017}, as a policy than can be learned using reinforcement learning \citep{Fang2017} or imitation learning \citep{Liu2018,Liu2018c,Vu2019}, and also as an adversarial problem \citep{Deng2018}.


\subsection{Prediction with Expert Advice}
\label{sec:experts}

A problem of prediction with expert advice can be described as an iterative game between a {\bf forecaster} and the {\bf environment}, in which the forecaster seeks advice from different sources ({\bf experts}) in order to predict what will happen in the environment \citep{cesa-bianchi06}. 
Thus, at each iteration $t$, the forecaster consults the predictions $\hat{p}_{j,t}, j = 1 \ldots J$ made by a set of $J$ weighted experts, in the decision space $\mathcal{D}$. 
Considering these predictions, the forecaster makes its own prediction, $\hat{p}_{f,t}\in\mathcal{D}$. At the same time, the environment reveals an outcome $y_t$ in the decision space $\mathcal{Y}$ (which may not necessarily be the same as $\mathcal{D}$). 
From this outcome, a loss can be derived in order to update the experts' weights $\omega_{1}, \ldots, \omega_{J}$. 

To learn the experts' weights, one can use an online learning algorithm, such as \ac{EWAF} \citep{cesa-bianchi06}, an algorithm with well-established performance guarantees. 
In \ac{EWAF}, the prediction made by the forecaster at each iteration $t$ is randomly selected following the probability distribution based on the experts' weights $\omega_{1,t-1} \ldots \omega_{J,t-1}$, as shown in Eq.~\ref{eq:PEWAF}:

\begin{equation}
\hat{p}_{f,t} = \frac{\sum_{j=1}^J\omega_{j,t-1} \hat{p}_{j,t}}{\sum_{j=1}^J\omega_{j,t-1}}
\label{eq:PEWAF}
\end{equation}

where $\hat{p}_{j,t}$ is the vector containing the current probabilities for each possible decision in $\mathcal{D}$, according to expert $j$. 
Then, the forecaster and each of the experts receive a non-negative loss ($\ell_{f,t}$ and $\ell_{j,t}$, respectively) based on the outcome $y_t$ revealed by the environment. 
%
%
The weight $\omega_{j,t}$ of each expert $j = 1 \ldots J$ is then updated according to the loss received by that expert, as follows:
\begin{equation}
\omega_{j,t}=\omega_{j,t-1}e^{-\eta\ell_{j,t}}
\label{eq:weightUpdate}
\end{equation} 

In the update rule above, if: 
\begin{equation}
\eta = \sqrt{\frac{8\log{J}}{T}}
\label{eq:etaUpdate}
\end{equation} 
it can be shown that the forecaster's {\bf regret} for not following the best expert's advice is bounded,  
as follows:
\begin{equation}
\sum_{t=1}^T\ell_{f,t} - \min_{j=1,\ldots,J} \sum_{t=1}^T \ell_{j,t} \leq\sqrt{\frac{T}{2}\log{J}}
\label{eq:EWAFregret}
\end{equation} 
i.e., that the forecaster quickly converges to the performance of the best expert after $T$ iterations \citep{cesa-bianchi06}.

\section{Related Work}
\label{sec:rw}

In this section, we present an overview of existing works that have applied the frameworks introduced in the previous section to \ac{MT}, namely active learning (Section~\ref{sec:ALxMT}), online learning (Section~\ref{sec:OLxMT}), and combinations of both (Section~\ref{sec:OLxAL}). 

\subsection{Active Learning for \acl{MT}}
\label{sec:ALxMT}

The earliest proposals that attempted to select training data in a clever way in \ac{MT}, despite not explicitly mentioning active learning, relied on criteria that could be seen as a query strategy (and ended up later inspiring active learning approaches), such as: unseen n-gram frequency and sentence TF-IDF \citep{Eck2005}, or similarity/n-gram overlap to the test set \citep{Hildebrand2005,Lu2007,Ittycheriah2007}. 
Since then, a vast variety of active learning approaches to \ac{MT} have been proposed, mostly in a pool-based\footnote{For a more comprehensive analysis of data selection and active learning approaches in a pool-based setting for \ac{MT} (up to 2015), see \citet{Eetemadi2015}.} and batch-mode settings. 

Focusing on Statistical \ac{MT}, query strategies proposed were mainly based on: phrase tables (e.g., phrase frequency on $\mathcal{L}$ or $\mathcal{U}$ \citep{Haffari2009}, phrase translation uncertainty/entropy \citep{Ambati2011}), language models (e.g., n-gram utility \citep{Haffari2009}, perplexity \citep{Mandal2008}, and KL divergence \citep{Ambati2012}), alignment of the parallel data \citep{Ambati2011}, the \ac{MT} model's confidence on the translation \citep{Haffari2009,Gonzalez-Rubio2011}, estimations of the translation error \citep{Ananthakrishnan2010} or of the translation's quality \citep{Logacheva2014}, and the round-trip translation accuracy (i.e., the error between a source sentence and the source obtained by translating the translation) \citep{Haffari2009}. 
Query strategies commonly seen in other tasks have also been employed in \ac{MT}, namely Query-by-Committee \citep{Mandal2008}, Information Density \citep{Gonzalez-Rubio2014}, Diversity, Density and combinations of both (e.g., static sentence sorting \citep{Eck2008}, Density-weighted Diversity \citep{Ambati2011}, n-gram coverage \citep{Gonzalez-Rubio2012,Gonzalez-Rubio2014}). 
All of these works addressed a pool-based setting, except for those of González-Rubio {\em et al} \citep{Gonzalez-Rubio2011,Gonzalez-Rubio2012,Gonzalez-Rubio2014}, who addressed a stream-based setting (although in the latter two works, the authors used pool-based strategies to select the most useful instances from the current batch, rather than using stream-based strategies). 

With the Neural \ac{MT} takeover, strategies based on neural models' details and Diversity/Density strategies based on embedding similarity have been preferred over n-gram, alignment, and phrase-based strategies. \citet{Peris2018} extended the proposals of González-Rubio et al for Statistical \ac{MT} \citep{Gonzalez-Rubio2011,Gonzalez-Rubio2012,Gonzalez-Rubio2014} by adding strategies based on coverage sampling (i.e. coverage of the attention weights over the source sentence, as a potential indicator of how good is the alignment between the source and the translation) and attention distraction (i.e., whether \ac{MT} model's attention is dispersed along the source sentence). 
\citet{Zhang2018} used the decoder's probability for the translation (as a form of uncertainty) and proposed a Diversity strategy based on subword-level embeddings, using {\sc fasttext} \citep{Bojanowski2017}. 
\citet{Zeng2019} made an extensive comparison of several query strategies found in the \ac{MT} literature on a Neural \ac{MT} system based on the Transformer architecture \citep{Vaswani2017}. They also introduced two new strategies: a variant of round-trip translation based on the likelihood of the source sentence according to the reverse translated model, and a Diversity strategy based on the cosine distance applied to the source sentences' subword-level embeddings (following \citet{Zhang2018}), contextual embeddings ({\sc Bert} \citep{Devlin2018}), and paraphrastic embeddings (i.e., contextual embeddings fine-tuned in a paraphrase dataset \citep{Wieting2018}). These two strategies, along with Density-weighted Diversity (following \citet{Ambati2011}), outperformed the remaining strategies in use. 
Finally, \citet{Hazra2021} addressed the problem of redundancy across the batch of instances selected by a given query strategy at each iteration, proposing the removal of redundant sentences using a model-aware similarity approach, on top of either one of three model-based query strategies: Uncertainty Sampling, coverage sampling, and attention distraction (following \citet{Peris2018} in the latter two strategies). 

Moreover, \acl{LTAL} strategies, which learn the query strategy from data, have also been applied to Neural \ac{MT}. \citet{Liu2018} viewed the query strategy as a policy network learned on a higher-resource language pair using imitation learning: the query strategy corresponds to a policy that learns by observing the inputs and output of an optimal policy; this optimal policy is trained on a higher-resource dataset, and its learning goal is to distinguish which instances are worthy to be annotated, based on how much each instance improves the performance of the task model on a development set, if added to the task model's labeled set. 
Their approach outperformed three heuristic strategies based on sentence length and uncertainty for most of the language pairs considered. 
\citet{Finkelstein2020} learned a a stream-based query strategy that should decide, for each incoming sentence, whether its translation should receive human feedback. This query strategy was a neural network pre-trained on a parallel corpus, which learned when to ask for feedback based on the {\sc chrF} score \citep{Popovic2015} between each sentence's automatic translation and its gold translation. 
The query strategy network was later retrained based on the human translator feedback (which can be given to a sentence when the query strategy asks for it, or by post-editing the final document), upon each document completion, based on the {\sc chrF} score between automatic and human made/post-edited translations. 
However, this proposal was not compared to other active learning query strategies.


Our proposal for applying active learning to \ac{MT} differs from these works in several aspects. First, only a few proposals consider a stream-based setting \citep{Gonzalez-Rubio2012,Gonzalez-Rubio2014,Peris2018}, and most of them end up applying pool-based strategies to select a subset of instances to be annotated among the current batch, instead of using stream-based strategies. 
Second, the two exceptions that use stream-based strategies rely solely on strategies that are either model-based \citep{Gonzalez-Rubio2011,Finkelstein2020} or require pre-training on additional data \citep{Finkelstein2020}. However, since our starting point is an ensemble of arbitrary (and potentially black-box) \ac{MT} models, and considering that we want to minimize the need for additional data, we constrain our solution to be model-independent and purely heuristic. 
%
Finally, our proposal differs from all of the works reviewed above in that we assume that the human-in-the-loop will not provide translations nor post-edits. Instead, we assume that the human will just rate the automatic translations, which is a form of feedback that requires less effort and is more plausible in practical applications such as Web \ac{MT} services or \ac{MT} shared tasks.


\subsection{Online Learning for \acl{MT}}
\label{sec:OLxMT}

There have been a number of online learning approaches applied to \ac{MT} in the past, mainly in Interactive \ac{MT} and/or post-editing \ac{MT} systems. 
Most approaches aimed at learning the parameters or feature weights of an \ac{MT} model \citep{Mathur2013,Denkowski2014,Ortiz-Martinez2016,Sokolov2016,Nguyen2017,Lam2018}, fine-tuning a pre-trained model for domain adaptation \citep{Turchi2017,Karimova2018,Peris2019}, or incorporating new parallel data from an unbounded stream \citep{Levenberg2010}.
 
Most of these works used human post-edited translations as a source of feedback, with the exceptions being the systems competing for \acs{WMT}'17 shared task on online bandit learning for \ac{MT} \citep{Sokolov2017}, as well as \citet{Lam2018}, who used (simulated) quality judgments.

In contrast to these approaches, few works used online learning to address the challenge of combining multiple \ac{MT} models and dynamically find the most appropriate ones for the language pair and set of models in hand. 
One of such works is that of \citet{Naradowsky2020}, who dynamically selected the best \ac{MT} system for a given \ac{MT} task or domain using stochastic multi-armed bandits and contextual bandits. The bandit algorithms learned from feedback simulated using a sentence-level {\sc Bleu} score \citep{papineni-bleu} between the selected automatic translation and a reference translation. 
Another one is a previous work of ours \citep{Mendonca2021}, in which we framed the problem of dynamically converging to the performance of the best individual \ac{MT} system as a problem of prediction with expert advice (when feedback is available to the translations outputted by all the systems in the ensemble) and adversarial multi-armed bandits \citep{Herbert1952,Lai1985} (when feedback is only available for the final translation chosen by the ensemble). We simulated the human-in-the-loop by using actual human ratings obtained from an \ac{MT} shared task \citep{Barrault2019}, when available in the data, and proposed different fallback strategies to cope with the lack of human ratings for some of the translations. 

We thus build upon our previous work \citep{Mendonca2021}, since it is the only one that takes advantage of human quality ratings rather than translations or post-edits. This setting not only reduces the human effort involved in improving the \ac{MT} ensemble, but it also should prove to be more suitable to represent real world \ac{MT} scenarios, such as Web translation systems or \ac{MT} shared tasks, in which the human-in-the-loop is not expected to provide a translation.


\subsection{Combining Online Learning and Active Learning}
\label{sec:OLxAL}

To the best of our knowledge, there are no works that combine multiple query strategies using online learning frameworks for \ac{MT}. The works that relate the most to our proposal in \ac{MT} are those of \citet{Haffari2009}, who combined multiple pool-based strategies in a weighted ensemble, learned {\em a priori} on a development set, and \citet{Peris2018}, who proposed an ensemble of stream-based query strategies, in which all strategies contributed equally (the strategies' votes are combined using a vote-entropy function \citep{dagan1995committee}). Both works differ from what we are proposing, since the weights of our ensemble of query strategies are learned dynamically based on human feedback, rather than fixed or learned {\em a priori}. 
Moreover, to the best of our knowledge, there are no online ensembles of stream-based query strategies in the literature either. 

However, the combination of multiple pool-based active learning strategies using online learning frameworks has been previously proposed for other tasks, mainly binary and multi-class classification. 

\citet{Baram2004} framed the problem of selecting a query strategy using multi-armed bandits \citep{Auer1995} and multi-armed bandits with experts \citep{Auer2002b}, which, unlike prediction with expert advice (recall Section~\ref{sec:experts}), assume that only the expert/arm selected knows its loss. The variant based on experts took advantage of multiple strategies and performed in line with the best individual strategy for binary classification problems. 

\citet{Osugi2005} also used multi-armed bandits with experts \citep{Auer2002b} to learn the best strategy for binary classification problems (but with a different loss function than \citet{Baram2004}), and outperformed the two individual query strategies considered, as well as \citet{Baram2004}. 

\citet{Hsu2015} presented a modification of an algorithm for multi-armed bandits with experts so that it would compute a reward (loss function) for all the query strategies that selected the same instance as the chosen strategy (instead of only to that strategy). This approach performed in line with the best individual query strategies, and outperformed \citet{Baram2004}. 

\citet{Chu2016} built on \citet{Hsu2015}, but applied contextual bandits instead, and also proposed to transfer the ensembles of query strategies learned in one dataset to other datasets. Their approach performed in line with the best two individual query strategies and outperformed \citet{Hsu2015}.

\citet{Pang2018} proposed a modification of multi-armed bandits with experts, to account for non-stationary loss functions (i.e., the best expert might vary over time), in binary and multi-class classification tasks. Their approach outperformed or performed in line with the best individual strategies and outperformed both \citet{Baram2004} and \citet{Hsu2015} in non-stationary datasets. 

Finally, in a previous work of ours \citep{Mendonca2020}, we combined multiple well-studied pool-based query strategies 
for two sequence labeling tasks (Part-of-Speech tagging and Named Entity Recognition), using prediction with expert advice. Our approach was able to converge to the performance of the best individual query strategies, and nearly reached the performance of a supervised baseline (which used twice the annotation budget). 

Inspired by our previous work, we propose to combine multiple query strategies using prediction with expert advice, now in a stream-based setting. Our goal is to converge to the most useful strategy without needing to evaluate them {\em a priori} for each language pair and set of \ac{MT} models, since a change in these factors might have an impact on which query strategy proves more useful, as discussed by \citet{Lowell2019}.

\section{Problem Definition}
\label{sec:problem}

\begin{figure}
\centering
\includegraphics[width=0.75\columnwidth]{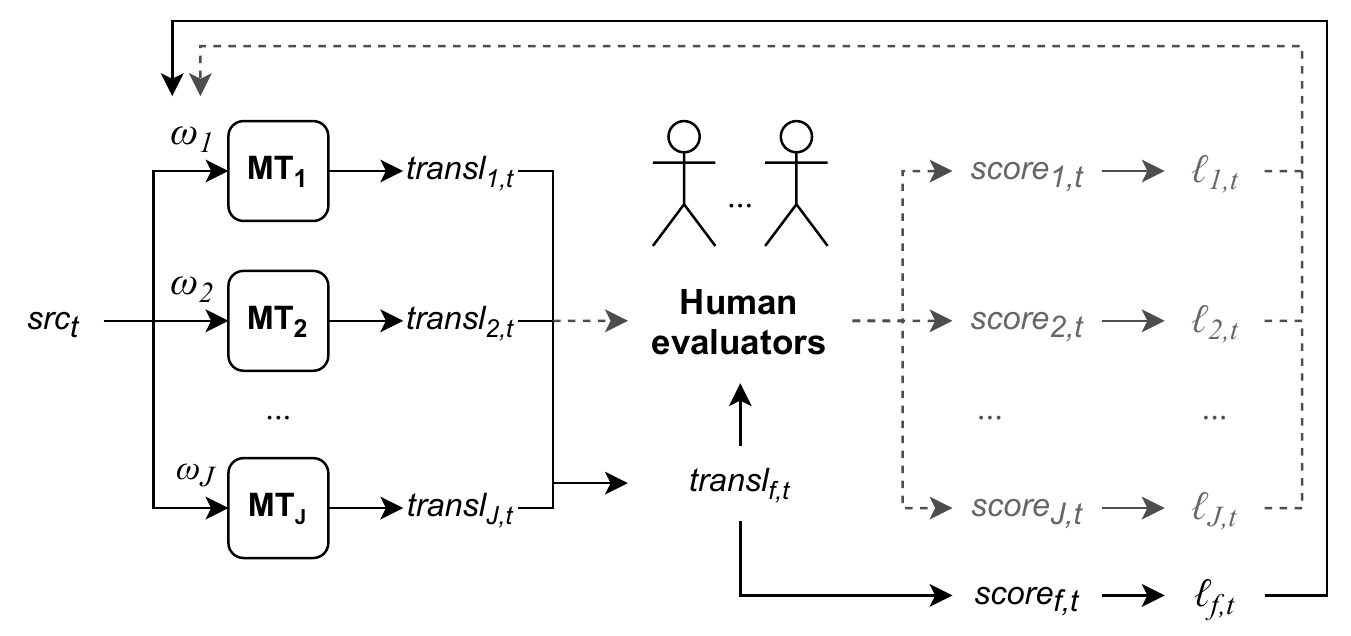}
\caption{Overview of the online learning process applied to \ac{MT}, at each iteration $t$. The grey dashed arrows represent flows that only occur when using prediction with expert advice \citep{Mendonca2021}.}
\label{fig:MTOnline}
\end{figure}

In this article, we build on our previous work in which we combined online learning and \ac{MT} \citep{Mendonca2021}, as described in Section~\ref{sec:OLxMT}. 
An overview of the online learning scenario is illustrated in Fig.~\ref{fig:MTOnline}: each \ac{MT} system is an expert (or arm) $j = 1 \ldots J$, associated with a weight $\omega_j$ (all the systems start with the same weight). At each iteration $t$, a segment $src_t$ is selected from the source language corpus and handed to all the \ac{MT} systems. Each system outputs a translation $transl_{j,t}$ in the target language, and one of these translations is selected as the forecaster's action according to the probability distribution given by the systems' weights. 
The chosen translation $transl_{f,t}$ (when using multi-armed bandits) or the translations outputted by all the systems (when using prediction with expert advice) receive a human assessment score $score_{j,t}$, from which a loss $\ell_{j,t}$ is derived for the respective \ac{MT} system. Finally, the weight $\omega_{f,t-1}$ of the chosen system (for the multi-armed bandits setting) or the weights $\omega_{1,t-1} \ldots \omega_{J,t-1}$ of all the systems (for the prediction with expert advice setting) are updated as a function of the loss received. 

In this work, our first challenge is to decide whether each incoming segment $src_t$ and respective translations $transl_{1,t}, \ldots, transl_{J,t}$ are informative enough so that the translations are worthy to be rated by a human. To address this challenge, we propose to use of a number of stream-based query strategies, which we detail in Section~\ref{sec:ALxOLxMT}. 

Our second challenge is to make the best use of multiple query strategies available, considering that we do not know {\em a priori} which query strategy will be the most adequate for a given language pair and set of models. To tackle this challenge, we propose to combine multiple strategies under the framework of prediction with expert advice, updating their weights in an online fashion. We describe this approach in Section~\ref{sec:OLxALxOLxMT}.

\section{Stream-based Active Learning for \ac{MT}}
\label{sec:ALxOLxMT}

\begin{figure}
\centering
\includegraphics[width=1\columnwidth]{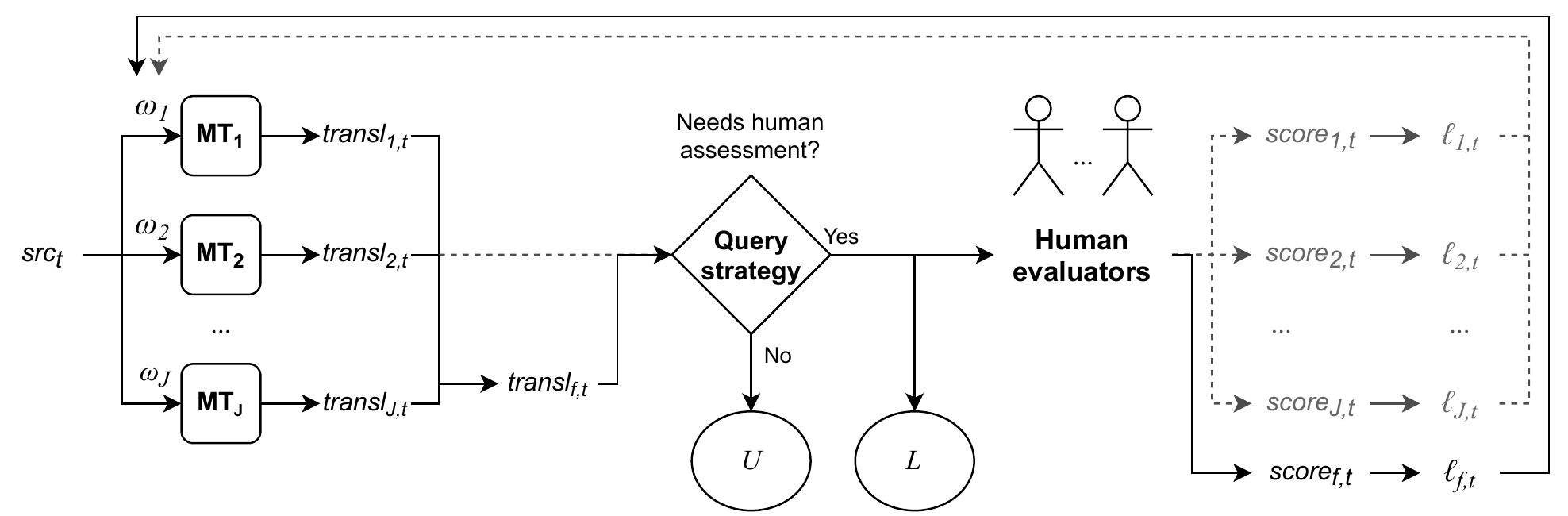}
\caption{Overview of the stream-based active learning process applied to our online \ac{MT} ensemble, at each iteration $t$. The grey dashed arrows represent flows that only occur when using prediction with expert advice on the \ac{MT} ensemble.}
\label{fig:ALxOLxMT}
\end{figure}

Given the online nature of our scenario, we operate under a stream-based setting (recall Section~\ref{sec:AL}), in which we assume that we only have access to one source segment (and its respective automatic translations) at a time. 
However, in order to more easily adapt certain query strategies to this setting, we store the segments that have been scored by the human in a human-scored set $\mathcal{L}$ and the segments that have been discarded (i.e., for which the human was not asked a score) in a discarded set $\mathcal{U}$. Since we are dealing with a stream-based setting, the decision of whether to select a certain segment is based on whether the values computed by each query strategy are above or below a given threshold. The stream-based active learning process applied to our \ac{MT} scenario is illustrated in Fig.~\ref{fig:ALxOLxMT}. 

\begin{algorithm}
\caption{Stream-based active learning for an online \acs{MT} scoring}
\label{alg:ALstreamMT}
\begin{algorithmic}[1]
    \REQUIRE \ac{MT} models $\mathcal{M} = m_1, \ldots, m_J$, stream of source segments $\mathcal{S}$, human-scored set $\mathcal{L}$, discarded set $\mathcal{U}$

    \STATE $\omega_{1,0}, \ldots, \omega_{J,0} \gets 1$

    \STATE $\mathcal{L}, \mathcal{U} \gets \emptyset $

    \FORALL{$src_t \in \mathcal{S}$}

        \FORALL{$m_j \in \mathcal{M}$} \label{l:gettrans} 
            \STATE $transl_{j, t} \gets m_j.translate(src_t)$ \COMMENT{Obtaining translations}
            \STATE $transl_t \gets transl_t  \cup transl_{j, t}$
        \ENDFOR \label{l:gettransend}
        
        \STATE $transl_{f, t}, m_f \gets forecasterMT(transl_t)$ \label{l:forecasterMT} \COMMENT{Choosing a translation}
        \IF[Query strategy]{$selectInstance?(src_t, transl_{f, t}, transl_t, \mathcal{L}, \mathcal{U})$} \label{l:qsstreamMT}
            \IF{prediction with expert advice} \label{l:ewafupdate}
                \FORALL{$m_j \in \mathcal{M}$}
                    \STATE $score_{j,t} \gets askHumanScore(src_t, transl_{j, t})$
                    \STATE $scores_t \gets scores_t  \cup score_{j, t}$
                    \STATE $\omega_{j, t} \gets m_j.updateWeight(score_{j,t})$ \COMMENT{\ac{MT} systems' weights update}
                \ENDFOR \label{l:ewafupdateend}

            \ELSIF{multi armed bandits} \label{l:exp3update}
                \STATE $score_{f,t} \gets askHumanScore(src_t, transl_{f, t})$
                \STATE $\omega_{f, t} \gets m_f.updateWeight(score_{f,t})$  \COMMENT{Chosen \ac{MT} system's weight update}
            \ENDIF \label{l:exp3updateend}

            \STATE $\mathcal{L} \gets \mathcal{L} \cup \{ src_t, transl_t, score_{f,t}, scores_t \}$ \label{l:addToL} \COMMENT{Tracking scored segments}
        \ELSE 
            \STATE $\mathcal{U} \gets \mathcal{U} \cup \{ src_t, transl_t \}$ \label{l:addToU} \COMMENT{Tracking segments that were not scored}
        \ENDIF

    \ENDFOR

\end{algorithmic}
\end{algorithm}

As illustrated in Algorithm~\ref{alg:ALstreamMT}, for each source segment  $src_t$, we obtain the translations outputted by each \ac{MT} model $m_j \in \mathcal{M}$ (lines~\ref{l:gettrans}-\ref{l:gettransend}). Given these translations, the forecaster chooses a translation $transl_{f, t}$ as the most likely to be the best (line~\ref{l:forecasterMT}), according to the online algorithm under consideration (\ac{EWAF} for prediction with expert advice, or \ac{EXP3} \citep{Auer1995} for multi-armed bandits). 
Then, we apply a query strategy to decide whether we should ask a human to score that segment's translations (line~\ref{l:qsstreamMT}). Depending on the query strategy in use, $\mathcal{L}$ or $\mathcal{U}$ may also be taken into account. 
If the query strategy decides that the segment's translations should be scored by the human, the weights $\omega_1, \ldots, \omega_J$ associated with the \ac{MT} models are updated based on the score received by the respective translations $transl_{1,t}, \ldots, transl_{J,t}$ (in the case of prediction with expert advice -- lines~\ref{l:ewafupdate}-\ref{l:ewafupdateend}), or only the weight of the \ac{MT} model that outputted the forecaster's choice, $m_f$, is updated (in the case of multi-armed bandits -- lines~\ref{l:exp3update}-\ref{l:exp3updateend}). 
The scored segment is then added to $\mathcal{L}$ (line~\ref{l:addToL}). 
If the current segment was not selected by the query strategy to be scored, it is added to $\mathcal{U}$ instead (line~\ref{l:addToU}). The process repeats itself for as long as the stream of segments goes on.

Our stream-based setting, as well as the model-independence and low-resource requirements in our scenario, severely constrain which query strategies are worthy to use. In other words, we want to {\em avoid} strategies that: (1): rely on model details, since we assume the \ac{MT} systems available to be black-box and that we only have access to the model's output (translation); (2) rely on an unlabeled set, since it is not available in a stream-based setting; (3) imply any kind of additional training (as it is the case, for instance, of \acl{LTAL} or strategies based on language models). 
We thus propose the use of the following criteria as query strategies: 

\begin{description}
\item[{\bf Translation Disagreement}:] We follow the \acl{QbC} strategy \citep{Seung1992}, by computing the disagreement among the translations outputted by the \ac{MT} models in the ensemble (see Eq.~\ref{eq:QbC}). If the average agreement among all the translations $transl_{1, t}, \ldots, transl_{J, t}$, $AvgAgr$, is below a given threshold, then the segment should be scored by the human. 
We compute the agreement between two translations using: (1) a lexical similarity measure (Jaccard \citep{Jaccard12similarityCoefficient}), considering the segment's words as the basic unit; (2) the cosine similarity between pre-trained contextual segment-level embeddings ({\sc Bert} \citep{Devlin2018}); (3) a translation evaluation metric ({\sc Bleu} \citep{papineni-bleu}), applied in a segment-level fashion. 
\begin{equation}
AvgAgr \left(transl_t\right) = \frac{1}{ \frac{J^2 - J}{2}} \left( {\sum_{j=1}^J \sum_{j'=1}^{j-1} agreement \left(transl_{j, t}, transl_{j', t} \right)} \right)
\label{eq:QbC}
\end{equation}

\item[{\bf Translation Difficulty}:] Inspired by earlier work that used \acl{QE} as a data selection criterion \citep{Logacheva2014}, we measure the difficulty of translating a segment $src_t$ using a \acl{QE} metric based on the perplexity of the translation, {\sc Prism} \citep{Thompson2020} (see Eq.~\ref{eq:PRISM}). If the average quality of a segment's translations, $AvgQuality$, is below a given threshold (i.e., if the metric does not expect the translations outputted by the \ac{MT} models considering the source segment $src_t$, then the segment should be scored by the human). 
\begin{equation}
AvgQuality \left(transl_t\right) = \frac{1}{J} \left( {\sum_{j=1}^J Prism \left(transl_{j, t}, src_{j, t} \right)} \right)
\label{eq:PRISM}
\end{equation}

\item[{\bf Diversity w.r.t. scored segments}:] We apply the Diversity strategy \citep{Fujii1998} to a stream-based setting by computing how much the current source segment $src_t$ differs from the human-scored set $\mathcal{L}$. If the average similarity between $src_t$ and each source segment in $\mathcal{L}$ is below a given threshold, then the segment should be scored by the human. We compute the similarity between two segments using: (1) a lexical similarity measure (Jaccard \citep{Jaccard12similarityCoefficient}), considering the segment's words as the basic unit, and (2) the cosine similarity between pre-trained contextual segment-level embeddings ({\sc Bert} \citep{Devlin2018}). We also compute a variant of Diversity based on n-gram coverage \citep{Ambati2011,Zeng2019}, according to Eq.~\ref{eq:ngramCoverageDiv} (where $\mathbb{I}$ is the indicator function).  
\begin{equation}
Div \left(src_t, \mathcal{L} \right) = \frac{ \sum_{s \in ngram\left( src_t \right)} \mathbb{I} \left( s \notin ngram\left( \mathcal{L} \right) \right) } { \vert ngram\left( src_t \right) \vert }
\label{eq:ngramCoverageDiv}
\end{equation}

\item[{\bf Density w.r.t. discarded segments}:] We introduce a modified version of Density \citep{Fujii1998}, in which we compare the current source segment $src_t$ to the segments in the discarded set $\mathcal{U}$ (i.e., those that were not scored by the human). In other words, if the average similarity between $src_t$ and each source segment in $\mathcal{U}$ is above a given threshold, then the segment should be scored by the human. We compute the similarity between two segments using the same measures as in the Diversity strategy. Once again, we also compute a variant based on n-gram coverage \citep{Ambati2011,Zeng2019}, according to Eq.~\ref{eq:ngramCoverageDen} (where $\#\left(s \vert \mathcal{X} \right)$ denotes the frequency of the n-gram $s$ in the n-grams of set $\mathcal{X}$, and $\lambda$ is a decay parameter to penalize n-grams that were seen in the human-scored set $\mathcal{L}$). 
\begin{equation}
Den\left(src_t, \mathcal{U}, \mathcal{L} \right) = \frac{ \sum_{s \in ngram\left( src_t \right)} \# \left(s \vert \mathcal{U} \right) e^{- \lambda \# \left(s \vert \mathcal{L} \right) }} {\vert ngram(src_t) \vert  \vert ngram(\mathcal{U}) \vert}
\label{eq:ngramCoverageDen}
\end{equation}
The inclusion of this strategy may be counter-intuitive, since it chooses segments that are representative of those ignored so far (i.e., of those which were considered not informative enough to be scored by the human); however, at a certain point, it might be the case that the ignored segments are actually more representative of potential future sentences; therefore, it may be useful to ask the human to score them. 
\end{description}

\section{Combining Query Strategies Using Expert Advice}
\label{sec:OLxALxOLxMT}

The introduction of active learning query strategies to decide whether to prompt the human to score the translation(s) outputted by the \ac{MT} ensemble raises a new challenge: choosing the most appropriate query strategy for a given language pair and \ac{MT} ensemble, without any prior knowledge nor pre-evaluation of the performance of each strategy. 
To tackle this challenge, we follow another previous work of ours, in which we combined different query strategies in a pool-based scenario for two sequence labeling tasks, using online learning under the framework of prediction with expert advice \citep{Mendonca2020}. 

\begin{figure}
\centering
\includegraphics[width=0.75\columnwidth]{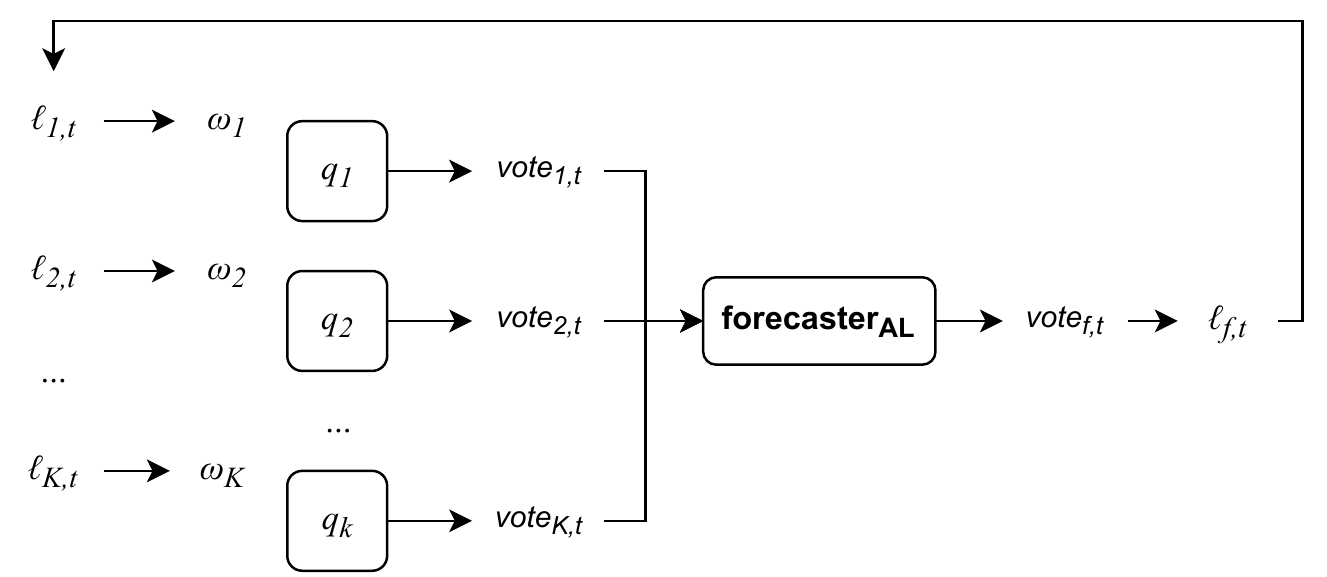}
\caption{Overview of the stream-based active learning process with expert advice at each iteration $t$.}
\label{fig:OLxAL}
\end{figure}

\begin{algorithm}
\caption{Stream-based active learning with expert advice}
\label{alg:onception}
\begin{algorithmic}[1]
    \REQUIRE \ac{MT} models $\mathcal{M} = m_1, \ldots, m_J$, stream of source segments $\mathcal{S}$, 
     query strategies  $\mathcal{Q} = q_1, \ldots, q_K$,  human-scored set $\mathcal{L}$, discarded set $\mathcal{U}$

    \STATE $\omega_{1,0}, \ldots, \omega_{J,0} \gets 1$
    \STATE $\omega_{1,0}, \ldots, \omega_{K,0} \gets 1$

    \STATE $\mathcal{L}, \mathcal{U} \gets \emptyset $

    \FORALL{$src_t \in \mathcal{S}$}
        \FORALL{$m_j \in \mathcal{M}$} 
            \STATE $transl_{j, t} \gets m_j.translate(src_t)$ \COMMENT{Obtaining translations}
            \STATE $transl_t \gets transl_t  \cup transl_{j, t}$
        \ENDFOR

        \STATE $transl_{f, t}, m_f \gets forecasterMT(transl_t)$ \COMMENT{Choosing a translation}

        \FORALL{$q_k \in \mathcal{Q}$} \label{l:getvoteAL}
            \STATE $vote_{k, t} \gets q_k.selectInstance?(src_t, transl_{f, t}, transl_t, \mathcal{L}, \mathcal{U})$ \COMMENT{Query strategy}
            \STATE $votes_t \gets votes_t \cup vote_{k, t}$
        \ENDFOR \label{l:getvoteALend}

        \STATE $vote_{f, t} \gets forecasterAL(votes_t)$ \label{l:forecasterAL} \COMMENT{Choosing a query strategy}

        \IF{$vote_{f, t} == True$}

            \IF{prediction with expert advice}
                \FORALL{$m_j \in \mathcal{M}$}
                    \STATE $score_{j,t} \gets askHumanScore(src_t, transl_{j, t})$
                    \STATE $scores_t \gets scores_t  \cup score_{j, t}$
                    \STATE $\omega_{j, t} \gets m_j.updateWeight(score_{j,t})$ \COMMENT{\ac{MT} systems' weights update}
                \ENDFOR

            \ELSIF{multi armed bandits}
                \STATE $score_{f,t} \gets askHumanScore(src_t, transl_{f, t})$
                \STATE $\omega_{f, t} \gets m_f.updateWeight(score_{f,t})$ \COMMENT{Chosen \ac{MT} system's weight update}
            \ENDIF 

            \STATE $R_{\mathcal{M}, t-1} \gets R_{\mathcal{M}, t}$ 
            \STATE $R_{\mathcal{M}, t} \gets updateRegret()$ \COMMENT{Updating the regret of the \ac{MT} ensemble}

            \FORALL{$q_k \in \mathcal{Q}$} 
                \STATE $\omega_{k, t} \gets q_k.updateWeight(R_{\mathcal{M}, t-1}, R_{\mathcal{M}, t})$ \label{l:updateEWALAL} \COMMENT{Query strategies' weights update}
            \ENDFOR 

            \STATE $\mathcal{L} \gets \mathcal{L} \cup \{ src_t, transl_t, score_{f,t}, scores_t \}$ \COMMENT{Tracking scored segments}
        \ELSE 
            \STATE $\mathcal{U} \gets \mathcal{U} \cup \{ src_t, transl_t \}$ \COMMENT{Tracking segments that were not scored}
        \ENDIF

    \ENDFOR

\end{algorithmic}
\end{algorithm}

In our stream-based active learning scenario, each expert corresponds to a query strategy $k = 1, \ldots, K$, to which a weight $\omega_k$ is assigned (all the query strategies start with the same weight). The learning process is illustrated in Fig.~\ref{fig:OLxAL} and Algorithm~\ref{alg:onception}. 
At each iteration, each query strategy casts a boolean value indicating whether the current segment's translation(s) should be scored by the human (lines~\ref{l:getvoteAL}-\ref{l:getvoteALend}). The forecaster randomly chooses one of the query strategies votes (line~\ref{l:forecasterAL}), based on the distribution of the strategies' weights (recall Eq.~\ref{eq:PEWAF} in Section~\ref{sec:experts}). Depending on the forecaster's vote, the current segment's translation(s) are scored by the human (or not), and the weights of the \ac{MT} systems in the ensemble are updated accordingly, as already seen in Section~\ref{sec:ALxOLxMT}. 

We want to award a greater weight to the strategies that casted the best votes, to make those strategies more likely to be considered in future iterations. This raises a question: how can we measure the success of each query strategy? Since our end goal is to improve the \ac{MT} ensemble, we measure its improvement at each weight update by considering the {\bf expected regret} $R_{\mathcal{M}}$ for not choosing the best \ac{MT} system's translation at each iteration, up to the current iteration $T$ (Eq.~\ref{eq:MTregret}), which can be seen as a {\em dynamic} regret \citep{Pang2018}. Note that this regret formulation deviates from the traditional formulation, in that we compare the forecaster to the best sequence of decisions overall (whose cumulative loss is given by $\sum_{t=1}^T \min_{j=1,\ldots,J} \ell_{j,t}$), instead of the best expert overall (whose cumulative loss would be given by $ \min_{j=1,\ldots,J} \sum_{t=1}^T \ell_{j,t}$). This way, we can ensure that our ensemble of active learning strategies learns from the best policy possible. 

\begin{equation}
R_{\mathcal{M}} = \sum_{t=1}^T\ell_{f,t}-\sum_{t=1}^T \min_{j=1,\ldots,J} \ell_{j,t}
\label{eq:MTregret}
\end{equation} 

Thus, at each iteration, we consider the variation of the expected regret, $\Delta_{R_{\mathcal{M}}} = R_{\mathcal{M}, t} - R_{\mathcal{M}, t-1}$
, to compute the loss function $\ell_{k,t}$ that, in turn, will allow to update the weights $\omega_1, \ldots, \omega_K$ of all the query strategies (line~\ref{l:updateEWALAL}): those who voted in favor of scoring will receive a loss proportional to the increase in the regret, while the remaining strategies will receive a loss inversely proportional to the regret's increase (Eq.~\ref{eq:regretloss}). 


\begin{equation}
\ell_{k,t} = 
\left\{
    \begin{array}{ll}
       \Delta_{R_{\mathcal{M}}}  & \mbox{ if } vote_{k, t} = 1  \\

        1 - \Delta_{R_{\mathcal{M}}} & \mbox{ if } vote_{k, t} = 0
    \end{array}
\right.
\label{eq:regretloss}
\end{equation} 

Note that, if the forecaster's vote was against human scoring, there is no change in the regret of the \ac{MT} ensemble, therefore we cannot update the weights of the query strategies. 

We follow a slightly different approach for the scenario when we only have partial feedback for the translations outputted by the \ac{MT} systems (i.e., when using multi-armed bandits to learn the weights of the \ac{MT} ensemble). Since only the \ac{MT} system corresponding to the currently chosen arm receives a loss, we do not know what would have been the optimal choice in hindsight, on a real world scenario. 
Thus, instead of considering the cumulative loss for the forecaster and what would be the optimal cumulative loss in hindsight, we compute the regret as the difference between the {\em average} loss for the forecaster and the {\em average} loss of the arm with the lowest average loss at that iteration (obtained considering the amount of times that arm was chosen). In this case, $\Delta_{R_{\mathcal{M}}}$ may vary between -1 and 1, thus we compute $\ell_{k,t}$ as shown in Eq.~\ref{eq:regretlossEXP3}. 

\begin{equation}
\ell_{k,t} = 
\left\{
    \begin{array}{ll}
        \frac{\Delta_{R_{\mathcal{M}}} + 1}{2}  & \mbox{ if } vote_{k, t} = 1  \\

        1 - \frac{\Delta_{R_{\mathcal{M}}} + 1}{2}      & \mbox{ if } vote_{k, t} = 0
    \end{array}
\right.
\label{eq:regretlossEXP3}
\end{equation} 

For the cases where the lowest loss is zero, not as a merit of the arm's performance, but because such arm has not been chosen yet, we assume its loss to be the average between zero and the highest average loss so far among the remaining arms.

\section{Experimental Setup}
\label{sec:setup}

To validate our proposals, we performed an experiment using data from an \ac{MT} shared task which includes human scores, allowing us to simulate an online setting. In this simulation, the goal of the online \ac{MT} ensemble is to give a greater weight to the top \ac{MT} systems competing for each language pair according to the shared task's official ranking, without knowing in advance which systems are the best.   
Our goal in this experiment is to converge to the top \ac{MT} systems with as little human intervention as possible. Thus, we want to observe how fast this happens: (1) using each individual stream-based query strategy proposed in Section~\ref{sec:ALxOLxMT}; (2) using the online ensemble of query strategies, proposed in Section~\ref{sec:OLxALxOLxMT}. 

In this section, we detail the data used (Section~\ref{sec:datasets}), the \ac{MT} loss functions (Section~\ref{sec:lossfunctions}), the implementation details for the query strategies (Section~\ref{sec:ALdetails}), the computing infrastructure on which the experiments were performed (Section~\ref{sec:compsetup}), and the evaluation metrics considered (Section~\ref{sec:evalmetrics}).


\subsection{Data}
\label{sec:datasets}

Following our previous work \citep{Mendonca2021}, we performed our experiments on the test datasets made available by the \acs{WMT}'19 News Translation shared task \citep{Barrault2019}, using the same selection of language pairs, since it offers a diverse coverage of phenomena in the dataset: 

\begin{itemize}
    \item English $\rightarrow$ German ({\tt en-de});
    \item French $\rightarrow$ German ({\tt fr-de});
    \item German $\rightarrow$ Czech ({\tt de-cs});
    \item Gujarati $\rightarrow$ English ({\tt gu-en});
    \item Lithuanian $\rightarrow$ English ({\tt lt-en}). 
\end{itemize}

The test sets for these language pairs are summarized in Table~\ref{tab:testsets}. For each language pair, each source segment is associated with the following information: 

\begin{itemize}
    \item A reference translation in the target language (produced specifically for the task);
    \item The automatic translation outputted by each system competing in the task for that language pair;
    \item The average score obtained by each automatic translation, according to human assessments made by one or more human evaluators, in two formats: a raw score in [0;100] and a z-score in [$-\infty;+\infty$]. Not all the automatic translations received a human assessment;
    \item The number of human evaluators for each automatic translation (if there were any).
\end{itemize}

\begin{table}[h]
\caption{Overview of the language pairs considered in our experiments.}
\label{tab:testsets}
\centering
\begin{tabular}{@{}rccccc@{}}
\toprule
\multicolumn{1}{l}{}       & \multicolumn{1}{c}{{\tt en-de}} & \multicolumn{1}{c}{{\tt fr-de}} & \multicolumn{1}{c}{{\tt de-cs}} & \multicolumn{1}{c}{{\tt gu-en}} & \multicolumn{1}{c}{{\tt lt-en}} \\ 
\midrule
Test set size (\# segments)              & 1,997                      & 1,701                      & 1,997                      & 1,016                      & 1,000                      \\
Competing systems          & 22                        & 10                        & 11                        & 12                        & 11                        \\
Human assessments coverage & 86.80\%                   & 23.52\%                   & 62.94\%                   & 75.00\%                   & 100.00\%                  \\ 
\bottomrule
\end{tabular}
\end{table}

For each language pair, we shuffled the respective test set once in our experiments, so that the original order of the segments would not bias the results. 


\subsection{\ac{MT} Loss Functions}
\label{sec:lossfunctions}

In order to compute the loss function of the online learning algorithms applied to the ensemble of \ac{MT} systems (\ac{EWAF} for prediction with expert advice, or \ac{EXP3} for multi-armed bandits), we used the human raw scores available in the test sets of the language pairs listed above. We normalized these  scores to be in the interval $[0,1]$ and rounded them to two decimal places, to avoid exploding weight values due to the exponential update rule. It should be noted that the use of these specific scores merely aims at simulating a human in a realistic situation, but our approach should be agnostic to the score format or range. 

As we can see in Table~\ref{tab:testsets}, not all segments received at least one human score in the shared task. Thus, we rely on the fallback strategies proposed on our previous work \citep{Mendonca2021}: 

\begin{itemize}
\item {\sc Human-Zero}: If there is no human score for the current translation, a score of zero is returned;
\item {\sc Human-Avg}: If there is no human score for the current translation, the average of the previous scores received by the system behind that translation is returned as the current score;
\item {\sc Human-Comet}: If there is no human score for the current translation, the {\sc Comet} score \citep{rei-etal-2020-comet} between the translation and the pair source/reference available in the corpus is returned as the current score (see \cite{Mendonca2021} for details).
\end{itemize}

In this experiment, for each combination of language pair and online algorithm, we considered only the fallback strategy that obtained the best results in \cite{Mendonca2021}, as shown in Table~\ref{tab:lossFunctions}.

\begin{table}[h]
\caption{Loss functions considered for each combination of language pair and online learning algorithm.}
\label{tab:lossFunctions}
\centering
\begin{tabular}{lll}
\toprule
\textbf{Language pair} & \textbf{EWAF} & \textbf{EXP3} \\
\midrule
{\tt en-de}                  & {\sc Human-Avg}      & {\sc Human-Zero}         \\
{\tt fr-de}                  & {\sc Human-Comet}    & {\sc Human-Zero}         \\
{\tt de-cs}                  & {\sc Human-Comet}    & {\sc Human-Comet}   \\
{\tt gu-en}                  & {\sc Human-Zero}     & {\sc Human-Comet}   \\
{\tt lt-en}                  & {\sc Human-Zero}     & {\sc Human-Zero}         \\
\bottomrule
\end{tabular}
\end{table}

It should be noted that, in our work, the decision on whether to prompt the human for feedback regarding a translation does not take into account whether the human is actually available, interested, or capable of providing such feedback (since, in a realistic scenario, one cannot guess that in advance). We thus keep these fallback strategies to cope with situations in which the human is prompted to provide feedback but chooses not do it. 


\subsection{Active Learning Implementation Details}
\label{sec:ALdetails}

For each combination of language pair and online learning algorithm applied to the ensemble of \ac{MT} models (\ac{EWAF} or \ac{EXP3}), we compared the following approaches:

\begin{itemize}
\item A baseline using all the data available to update the weights of the \ac{MT} models;
\item Each individual query strategy proposed in Section~\ref{sec:ALxOLxMT} and summarized below, plus a {\bf random} strategy:

    \begin{itemize}
        \item Diversity based on (1) Jaccard ({\bf DivJac}), and (2) the cosine of sentence-level {\sc Bert} ({\bf DivBERT});
        \item Density based on (1) Jaccard ({\bf DenJac}), and (2) the cosine of sentence-level {\sc Bert} ({\bf DenBERT});
        \item Translation Disagreement based on (1) Jaccard ({\bf TDisJac}), (2) the cosine of sentence-level {\sc Bert} ({\bf TDisBERT}), and (3) sentence-level {\sc Bleu} ({\bf TDisBLEU}); 
        \item Translation Difficulty based on {\sc Prism} ({\bf TDiff});
        \item N-gram coverage Diversity ({\bf DivNgram}) and Density ({\bf DenNgram}).
    \end{itemize}
\item An active learning approach combining multiple individual query strategies using prediction with expert advice, as described in Section~\ref{sec:OLxALxOLxMT}:
    
    \begin{itemize}
        \item {\bf Onception (all)} combines all the query strategies listed above (including the random baseline);
        \item {\bf Onception (no Density)} combines all the query strategies except the Density based ones, since this query strategy is based on segments that were previously discarded for scoring;
        \item {\bf Onception (no Density and TDiff)} combines all the query strategies except the Density based ones and Translation Difficulty; we only report this variation for {\tt gu-en} since {\sc Prism} was not trained in Gujarati.
    \end{itemize} 
\end{itemize}

Concerning the implementation of the query strategies, we made the following decisions. For the strategies based on n-gram coverage, we extracted all n-grams up to trigrams. 
For Translation Disagreement, we computed {\sc Bleu} using SacreBLEU \citep{Post2018}. 
For the query strategies using contextual embeddings, we extracted pre-trained embeddings using {\sc Bert} as a Service \citep{xiao2018bertservice} and the Base Cased models\footnote{\url{https://github.com/google-research/bert/}} (English for the English segments and Multilingual for the remaining languages). Before extracting the embeddings, as well as when using Jaccard in the Diversity and Density strategies, we removed punctuation from the segments. For n-gram coverage strategies, we also lowercased the segments before extracting n-grams. 

As for the thresholds used to decide whether to prompt the human for feedback, we performed a number of ablation runs with a range of thresholds. For each strategy, we only report the results obtained with the threshold that provided the best trade-off between performance and reducing human interaction (refer to Appendix A for the list of thresholds used). 


\subsection{Computing Infrastructure}
\label{sec:compsetup}

We performed our experiment on a PC with an Intel Core CPU i7-9750H @ 2.60GHz CPU, 32GB RAM, and a NVIDIA GeForce RTX 2060 6GB GDDR6, except for the extraction of the {\sc Bert} embeddings, which was performed on a shared machine with 24 Intel Xeon CPU E5-2630 v2 @ 2.60GHz and 257 GB RAM.


\subsection{Evaluation Metrics}
\label{sec:evalmetrics}

The main goal of our experiment is to observe how fast each approach converges (i.e., gives a greater weight) to the best \ac{MT} models for each language pair. As the gold standard, we consider the top ranked models on \acs{WMT}'19, listed in Table~\ref{tab:top3systems}. Thus, we report the performance of each approach by computing the overlap between the top $n=3$ systems with greatest weights according to our approaches, $\hat{s_n}$, and the top $n=3$ systems according to the shared task's official ranking, $s_n^*$:

\begin{equation}
OverlapTop_n = \frac{ \vert \hat{s_n} \cap s_n^* \vert }{n} , n = 3
\label{eq:tops}
\end{equation}

\begin{table}[h]
\caption{Top 3 performing systems for each language pair in the \acs{WMT}'19 News Translation shared task \citep{Barrault2019}. The systems named ``online-[letter]'' correspond to publicly available translation services and were anonymized in the shared task.}
\label{tab:top3systems}
\centering
\begin{tabular}{@{}clrr@{}}
\toprule
\multicolumn{1}{l}{}            & \textbf{Top 3}      & \multicolumn{1}{l}{\textbf{z-score}} & \multicolumn{1}{l}{\textbf{Raw score}} \\ 
\midrule

\multirow{3}{*}{\STAB{\rotatebox[origin=c]{90}{\texttt{en-de}}}} 
                                & Facebook-FAIR \citep{Ng2019}       & 0.347                                & 90.3                                   \\
                                & Microsoft-sent-doc \citep{Junczys-Dowmunt2019}  & 0.311                                & 93.0                                   \\
                                & Microsoft-doc-level \citep{Junczys-Dowmunt2019} & 0.296                                & 92.6                                   \\ \cmidrule{1-4}
\multirow{3}{*}{\STAB{\rotatebox[origin=c]{90}{\texttt{fr-de}}}} 
                                & MSRA-MADL \citep{Xia2019}           & 0.267                                & 82.4                                   \\
                                & eTranslation \citep{Oravecz2019}       & 0.246                                & 81.5                                   \\
                                & LIUM \citep{Bougares2019}               & 0.082                                & 78.5                                   \\ \cmidrule{1-4}
\multirow{3}{*}{\STAB{\rotatebox[origin=c]{90}{\texttt{de-cs}}}} 
                                & online-Y            & 0.426                                & 63.9                                   \\
                                & online-B            & 0.386                                & 62.7                                   \\
                                & NICT \citep{Dabre2019}               & 0.367                                & 61.4                                   \\ \cmidrule{1-4}
\multirow{3}{*}{\STAB{\rotatebox[origin=c]{90}{\texttt{gu-en}}}} 
                                & NEU \citep{Li2019}                & 0.210                                & 64.8                                   \\
                                & UEDIN \citep{Bawden2019}              & 0.126                                & 61.7                                   \\
                                & GTCOM-Primary \citep{Bei2019}      & 0.100                                & 59.4                                   \\ \midrule
\multirow{3}{*}{\STAB{\rotatebox[origin=c]{90}{\texttt{lt-en}}}} 
                                & GTCOM-Primary \citep{Bei2019}      & 0.234                                & 77.4                                   \\
                                & tilde-nc-nmt \citep{Pinnis2019}       & 0.216                                & 77.5                                   \\
                                & NEU \citep{Li2019}                & 0.213                                & 77.0                                   \\ 
\bottomrule
\end{tabular}
\end{table}

We are focused only on the top \ac{MT} systems since, in a realistic scenario (e.g., a Web MT service), a user would most likely only care about the main translation returned, or would at most consider one or two alternative translations. 
Moreover, the success of our approach is not restricted to converging to the best system, since the scores obtained in the shared task are not reliable enough to discriminate between similarly performing systems (due to the lack of a large enough coverage of human assessments). 
However, for a more detailed account of the performance of each approach, we report additional metrics in the appendix: 
(1) the evolution of the rank correlation (using Kendall's $\tau$) between the \ac{MT} systems sorted by their online learning weights and the official ranking (Appendix B); 
(2) the evolution of the weights of the query strategies when using Onception (Appendix C). 

\section{Results and Discussion}
\label{sec:results}

\begin{figure}
\centering
\includegraphics[width=1\columnwidth]{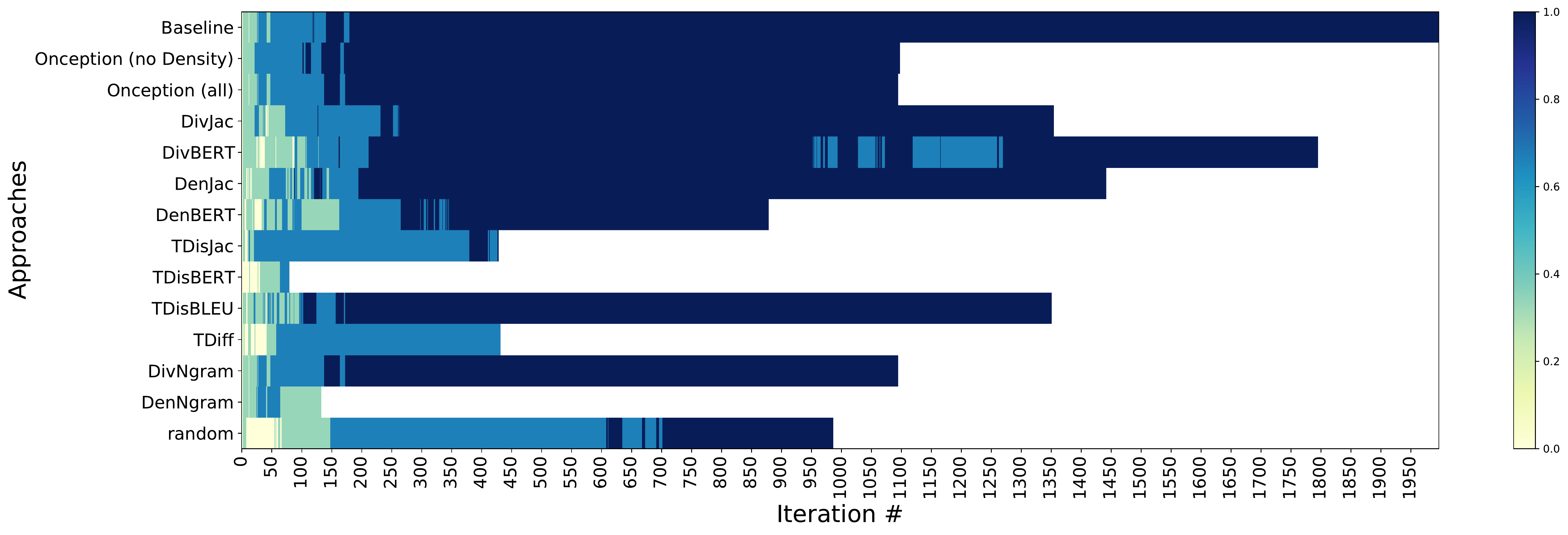}
\caption{Overlap of the top 3 \ac{MT} systems for {\tt en-de} and \ac{EWAF} (figure best seen in color).}
\label{fig:overlapEnDeEWAF}
\end{figure}

\begin{figure}
\centering
\includegraphics[width=1\columnwidth]{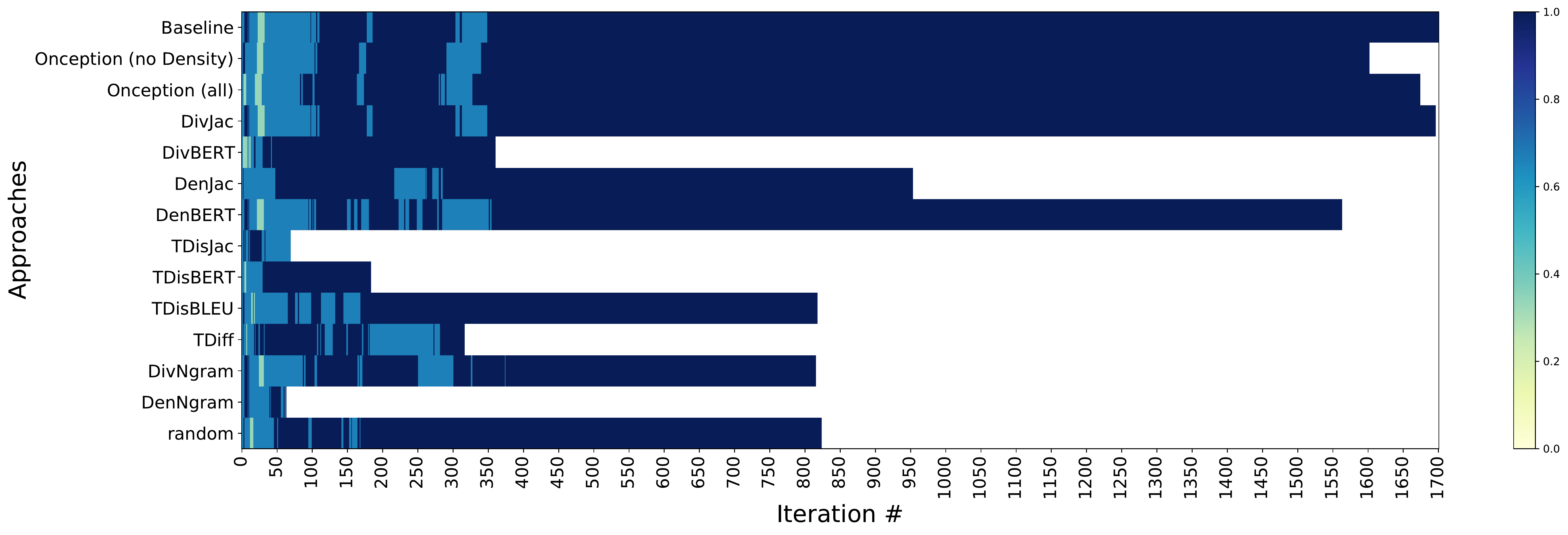}
\caption{Overlap of the top 3 \ac{MT} systems for {\tt fr-de} and \ac{EWAF} (figure best seen in color).}
\label{fig:overlapFrDeEWAF}
\end{figure}

\begin{figure}
\centering
\includegraphics[width=1\columnwidth]{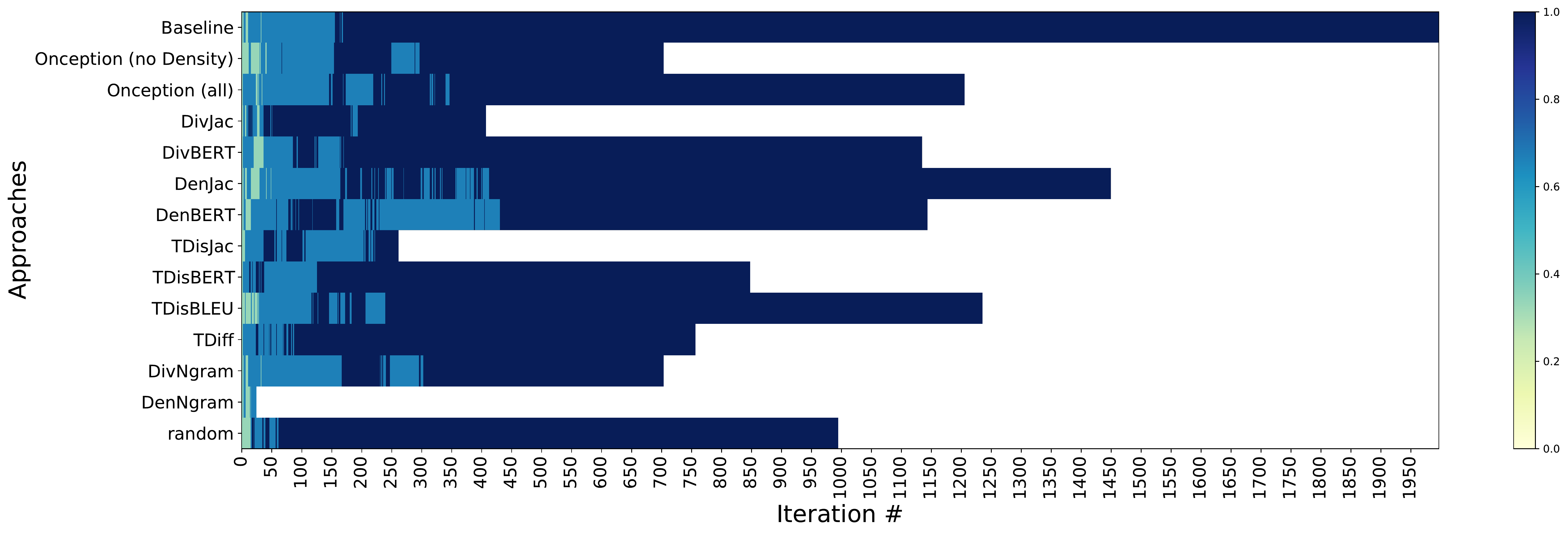}
\caption{Overlap of the top 3 \ac{MT} systems for {\tt de-cs} and \ac{EWAF} (figure best seen in color).}
\label{fig:overlapDeCsEWAF}
\end{figure}

\begin{figure}
\centering
\includegraphics[width=1\columnwidth]{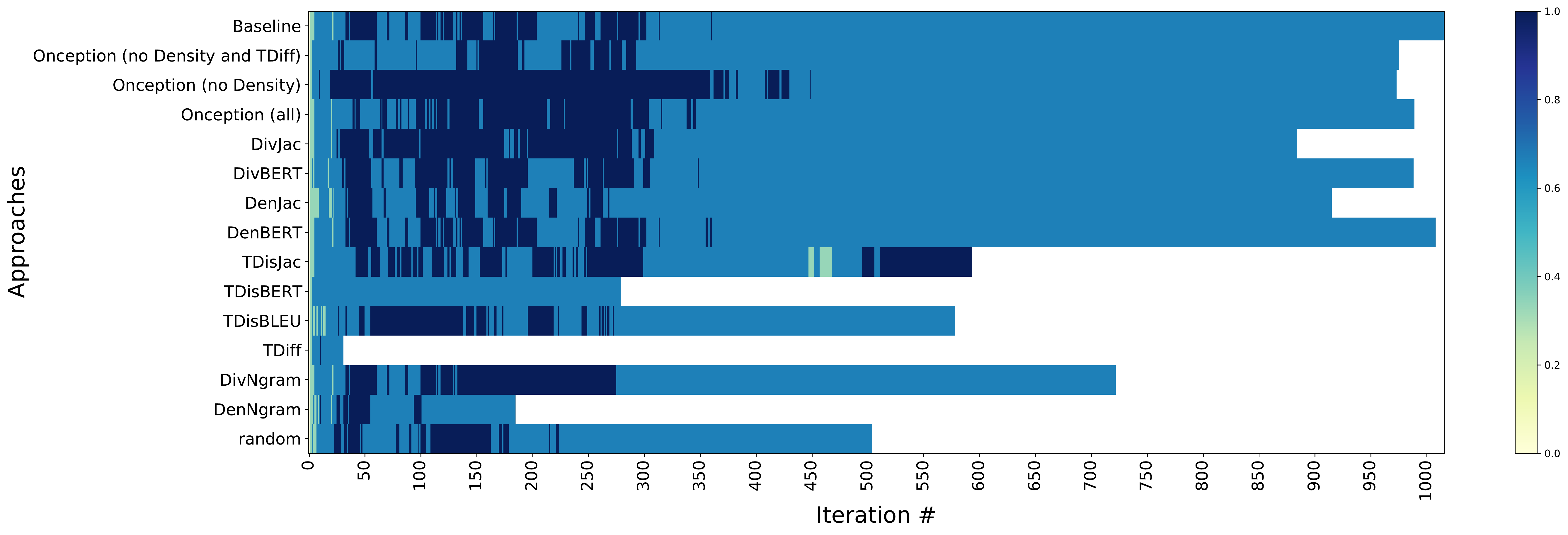}
\caption{Overlap of the top 3 \ac{MT} systems for {\tt gu-en} and \ac{EWAF} (figure best seen in color).}
\label{fig:overlapGuEnEWAF}
\end{figure}

\begin{figure}
\centering
\includegraphics[width=1\columnwidth]{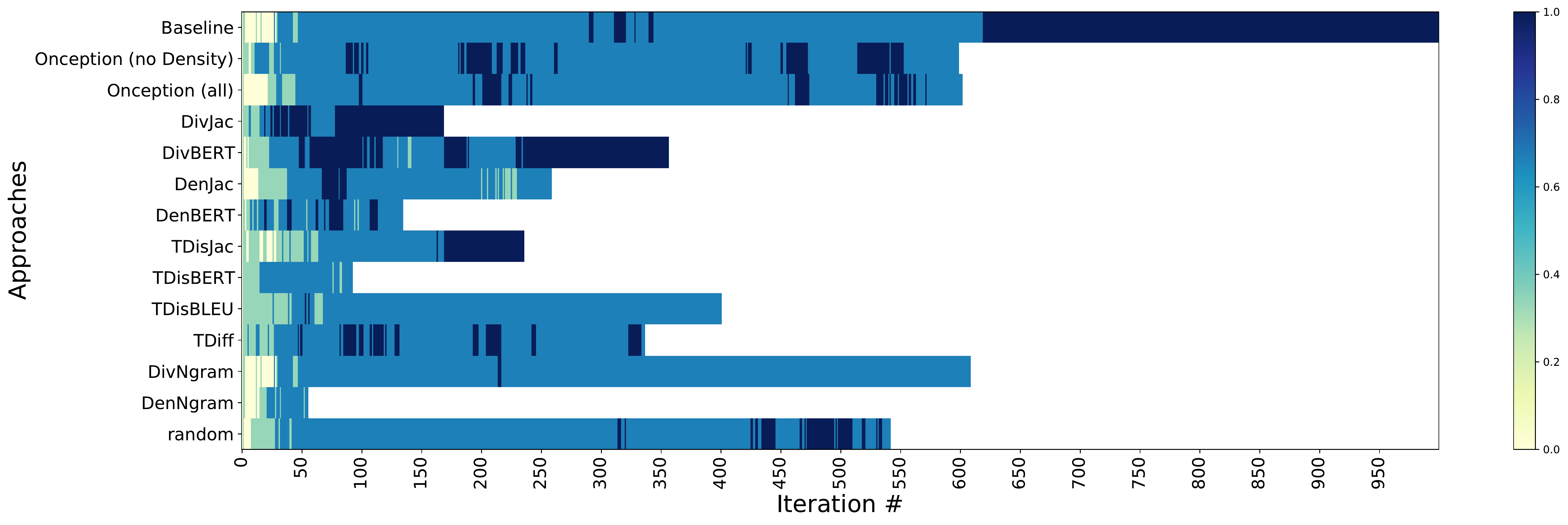}
\caption{Overlap of the top 3 \ac{MT} systems for {\tt lt-en} and \ac{EWAF} (figure best seen in color).}
\label{fig:overlapLtEnEWAF}
\end{figure}

Figs.~\ref{fig:overlapEnDeEWAF}-\ref{fig:overlapLtEnEWAF} represent the evolution of the $OverlapTop_{n=3}$ for the scenario in which all the translations for a given segment receive a human (or fallback) score (i.e., when using \ac{EWAF} to learn the selected \ac{MT} system's weight). Each iteration corresponds to a segment that was seen and whose translations were scored. In other words, the approaches for which the heatmap bars are shorter consult the human fewer times, thus making fewer updates to the \ac{MT} systems' weights.

The first thing we can notice is that, for all language pairs, most active learning approaches allow to reduce the amount of segments for which the human is prompted to score the translations (as indicated by the shorter heatmap bars). 
For {\tt en-de}, most active learning approaches follow an evolution similar to that of the baseline, with some of them (Onception (no Density), DenJac, and TDisBLEU) reaching a full overlap of the top 3 systems when the baseline does not. 
As for {\tt fr-de}, both Onception and Onception (no Density) converge to the full top 3 systems slightly earlier than the baseline, and some individual query strategies do so even earlier and/or when the baseline does not, namely: DivBERT, DenJac, TDisJac, TDisBERT, TDisBLEU, TDiff, and Random.
For {\tt de-cs}, both Onception and Onception (no Density) find the full top 3 systems earlier than the baseline (albeit not doing so consistently). Once again, several query strategies do so even earlier, namely: DivJac, DivBERT, DenBERT, TDisJac, TDisBERT, TDisBLEU, TDiff, and Random. 
As for {\tt gu-en}, Onception (no Density) performs better overall than both the baseline and the individual query strategies, Moreover, several query strategies reach the full top 3 when the baseline does not, most notably: DivJac, TDisJac, TDisBLEU, and DivNgram. 
Surprisingly, the variant of Onception that does {\em not} consider TDiff performs generally worse than most strategies, even though the quality estimation metric used ({\sc Prism}) was not trained on Gujarati. 
Finally, for {\tt lt-en}, the baseline takes over 600 updates to stabilize on the full top 3, but several query strategies manage to get there (albeit temporarily in some cases) much earlier, namely: Onception, Onception (no Density), DivJac, DivBERT, TDisJac, TDiff, and even Random. 

\begin{figure}
\centering
\includegraphics[width=1\columnwidth]{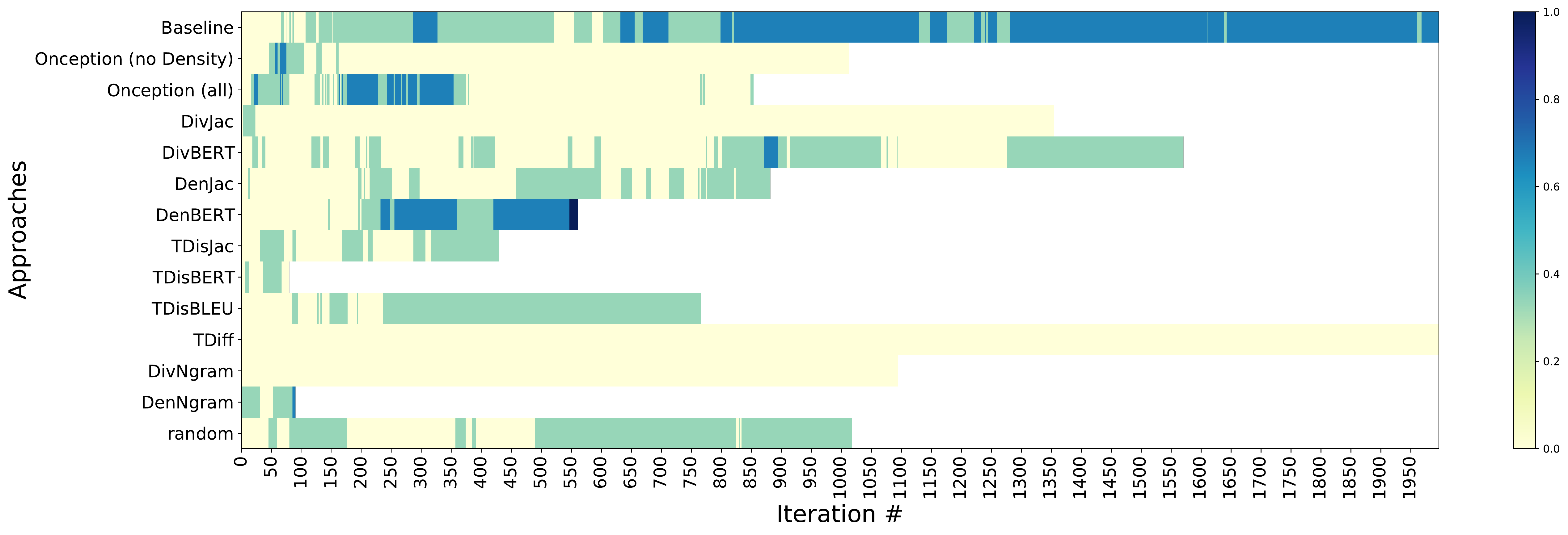}
\caption{Overlap of the top 3 \ac{MT} systems for {\tt en-de} and \ac{EXP3} (figure best seen in color).}
\label{fig:overlapEnDeEXP3}
\end{figure}

\begin{figure}
\centering
\includegraphics[width=1\columnwidth]{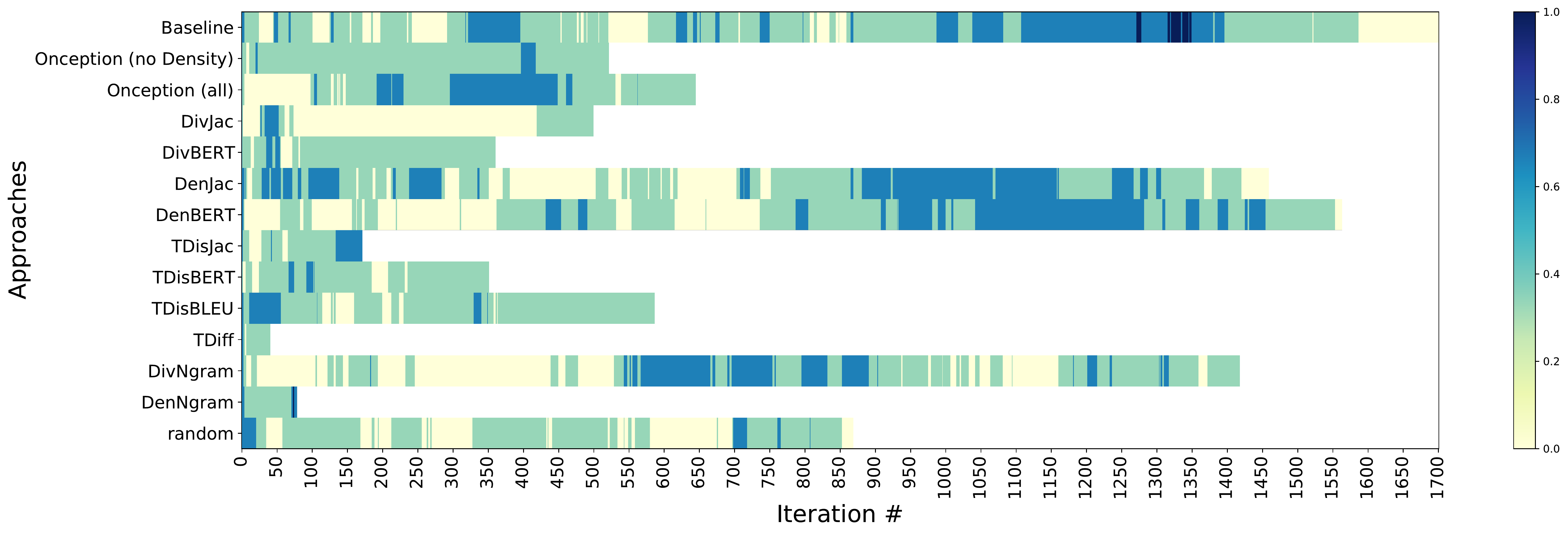}
\caption{Overlap of the top 3 \ac{MT} systems for {\tt fr-de} and \ac{EXP3} (figure best seen in color).}
\label{fig:overlapFrDeEXP3}
\end{figure}

\begin{figure}
\centering
\includegraphics[width=1\columnwidth]{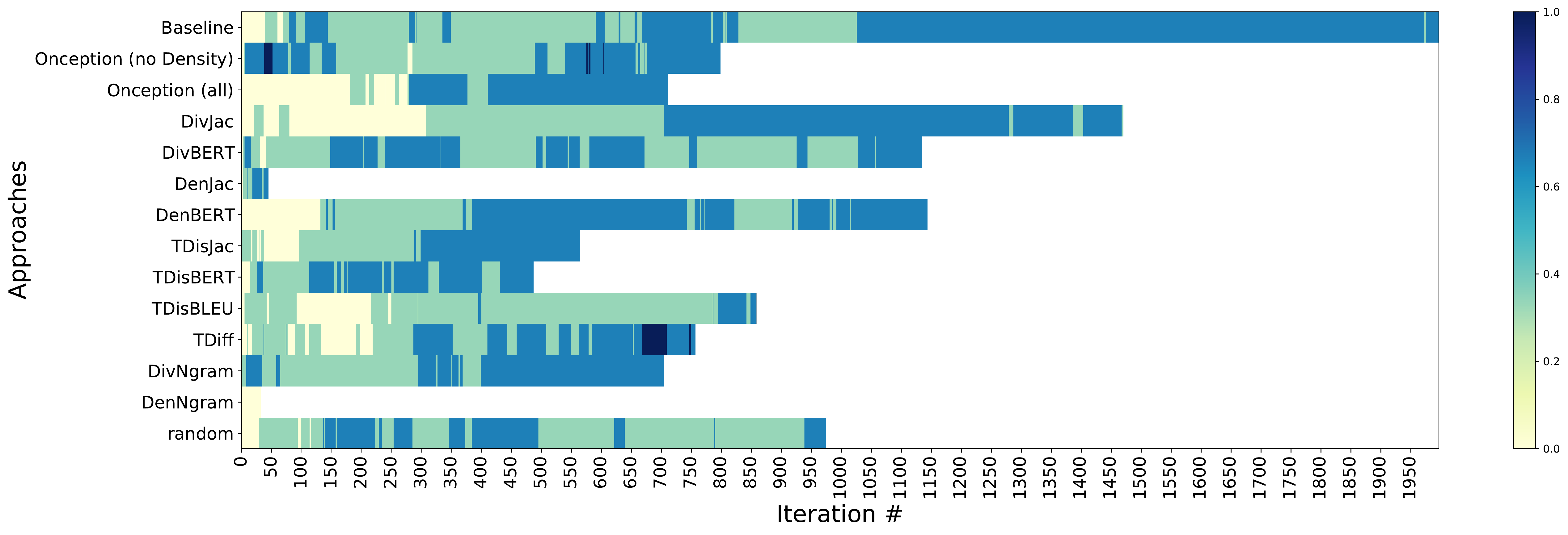}
\caption{Overlap of the top 3 \ac{MT} systems for {\tt de-cs} and \ac{EXP3} (figure best seen in color).}
\label{fig:overlapDeCsEXP3}
\end{figure}

\begin{figure}
\centering
\includegraphics[width=1\columnwidth]{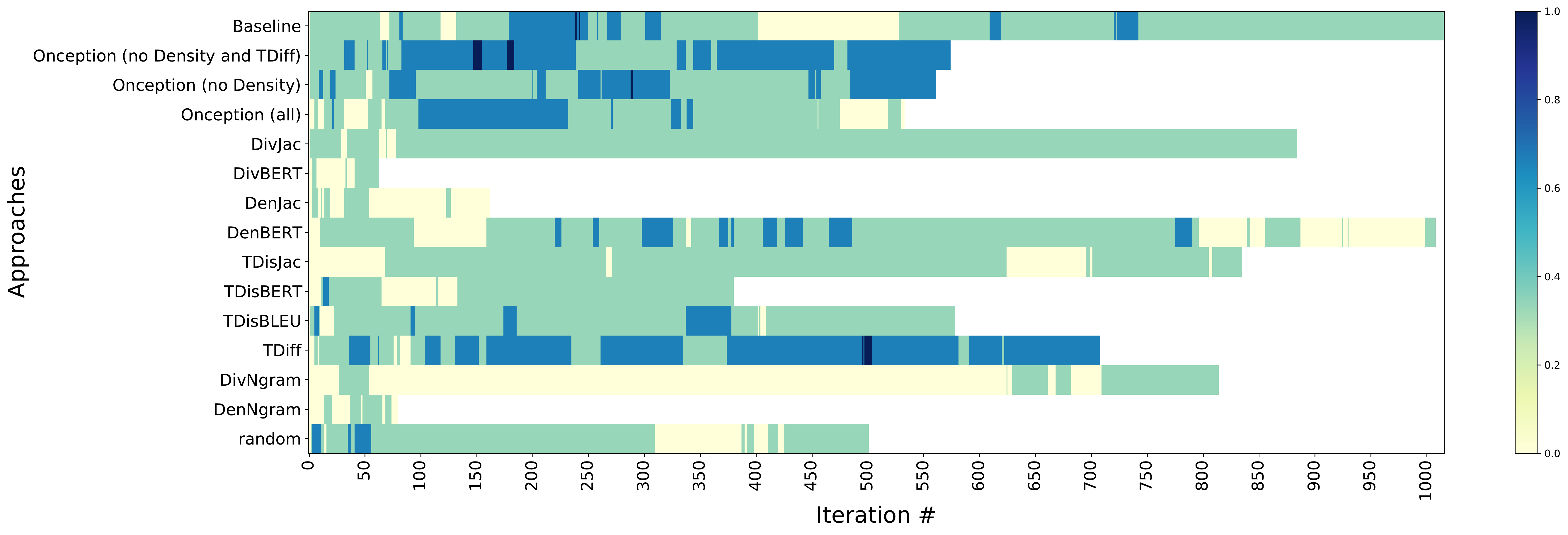}
\caption{Overlap of the top 3 \ac{MT} systems for {\tt gu-en} and \ac{EXP3} (figure best seen in color).}
\label{fig:overlapGuEnEXP3}
\end{figure}

\begin{figure}
\centering
\includegraphics[width=1\columnwidth]{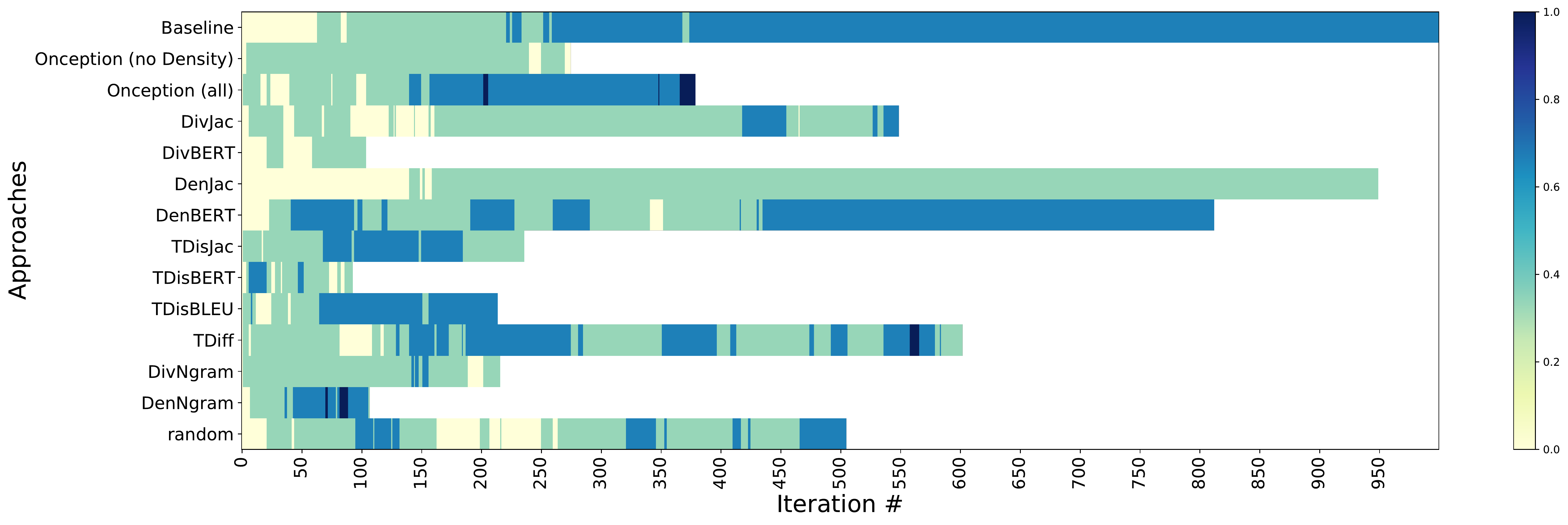}
\caption{Overlap of the top 3 \ac{MT} systems for {\tt lt-en} and \ac{EXP3} (figure best seen in color).}
\label{fig:overlapLtEnEXP3}
\end{figure}

Figs.~\ref{fig:overlapEnDeEXP3}-\ref{fig:overlapLtEnEXP3} represent the evolution of the $OverlapTop_{n=3}$ for the the scenario in which only the forecaster's selected translation receives a human (or fallback) score (i.e., when using \ac{EXP3} to learn the \ac{MT} systems' weights). 
Once again, each iteration corresponds to a segment that was seen and whose translation were scored (i.e., to an update to the weight of the selected \ac{MT} system). 
Since this scenario assumes fewer feedback, the overall performance of both the baseline and all the active learning approaches is generally more inconsistent than in the \ac{EWAF} scenario. 

Starting with {\tt en-de}, while the baseline overall performs better, both Onception and Onception (no Density), as well as DenBERT, are able to find the full top 3, while the baseline never does. 
For {\tt fr-de}, all approaches perform inconsistently, often alternating between never finding the top 3 systems and getting an overlap of 2 systems out of 3. Only the baseline (after around 1300 updates) and DenNgram (after fewer than 100 updates) get the full top 3 right. However, Onception (no Density) gets at least one top system right more consistently than both the baseline and most individual strategies. 
For {\tt de-cs}, most approaches generally find at least one of the 3 top systems. Only Onception (no Density) and TDiff find the full top 3, with Onception (no Density) doing so earlier (around 50 updates) and performing better overall during the first 150 updates. 
As for {\tt gu-en}, most approaches find at least one of the 3 top systems. Only Onception (no Density and TDiff), Onception (no Density), and TDiff find the full top 3, performing generally better than the remaining approaches, including the baseline. 
Finally, for {\tt lt-en}, most active learning approaches converge to a larger subset of the top 3 systems with fewer updates than the baseline, with only three of them (rarely) finding the full top 3: Onception (at around 200 and 375 updates) TDiff (at about 500 updates) and DenNgram (between around 50 to 100 updates). 

Following, we will see how these results answer our research questions. 

\subsection*{RQ1: Can an active learning approach find the best systems with fewer human interactions?}
\label{sec:RQAL}

As observed in the previous section, for all language pairs and online algorithms considered, there is at least one  query strategy that accelerates the process of finding the top 3 systems (or a subset of them). However, {\em which} query strategy should be used in each situation is not a trivial question to answer, since the performance of the query strategies under consideration varies greatly across language pairs and online algorithms.


\subsection*{RQ2: Does it pay off to combine multiple query strategies using expert advice (rather than using a single strategy)?} 
\label{sec:RQonception}

Considering the result discussed in the previous research question (i.e., that the best performing query strategy varies across settings), it generally pays off combining multiple strategies using online learning (Onception), since we do not know in advance which strategy will perform better for a given setting, and could end up committing to one of the worst ones. 
By looking at the heatmaps in Figs.~\ref{fig:overlapEnDeEWAF}-\ref{fig:overlapLtEnEXP3}, one could think that Onception would not always pay off, since this approach is not always well aligned with the performance of the best query strategies (i.e., it does not verify the performance guarantees of \ac{EWAF}).  
However, this behavior is expectable since Onception is learning from an {\em imperfect} source of feedback -- the expected regret of the \ac{MT} ensemble, which, during the first updates, may not reflect accurately what would be the best options overall. Even so, our Onception approaches often perform better than the baseline and most individual strategies, finding the full top 3 systems earlier (i.e., requiring less human effort).

\section{Conclusions and future work}
\label{sec:cfw}

In this work, we contributed with a active learning solution that reduces the human effort needed to learn an online ensemble of \ac{MT} models. We adapted multiple well-known query strategies to a stream-based setting, all of them being model-independent and compliant with a low-resource setting (as they do not require any additional pre-training data). Since we do not know in advance which query strategies will be more appropriate on a given setting, we dynamically combined multiple strategies using prediction with expert advice. This revealed itself to be a safer option than to commit to a single query strategy, often outperforming several individual strategies and sparing the need for human interaction even further. 

For future work, some extensions to our solution could be considered, namely: using dynamic thresholds for the query strategies, rather than fixed ones (since some strategies may need to be more strict at some points of the learning process); using stochastic and/or contextual bandits, as well as non-stationary experts, to combine the query strategies (since the best strategy tends to vary as new segments arrive, and a greater emphasis on exploration might be beneficial); learning the best \ac{MT} systems from more reliable ratings, such as \ac{MQM} based ratings \citep{mqm} (for instance, by replicating the experiments on the \acs{WMT}'21 datasets \citep{akhbardeh-etal-2021-findings}).


\appendix

\begin{landscape}

\appendixsection{Query Strategies' Thresholds}
\label{app:thresholds}

\begin{table}[h]
\caption{Threshold values used for each query strategy, language pair and online learning algorithm.}
\label{tab:thresholds}
\centering
\begin{tabular}{lrrrrrrrrrr}
\hline\noalign{\smallskip}
\multicolumn{1}{c}{\textbf{Strategy}} & \multicolumn{2}{c}{\textbf{en-de}}                                    & \multicolumn{2}{c}{\textbf{fr-de}}                                    & \multicolumn{2}{c}{\textbf{de-cs}}                                    & \multicolumn{2}{c}{\textbf{gu-en}}                                    & \multicolumn{2}{c}{\textbf{lt-en}}                                    \\
\noalign{\smallskip}\hline\noalign{\smallskip}
\textbf{}                             & \multicolumn{1}{c}{\textbf{EWAF}} & \multicolumn{1}{c}{\textbf{EXP3}} & \multicolumn{1}{c}{\textbf{EWAF}} & \multicolumn{1}{c}{\textbf{EXP3}} & \multicolumn{1}{c}{\textbf{EWAF}} & \multicolumn{1}{c}{\textbf{EXP3}} & \multicolumn{1}{c}{\textbf{EWAF}} & \multicolumn{1}{c}{\textbf{EXP3}} & \multicolumn{1}{c}{\textbf{EWAF}} & \multicolumn{1}{c}{\textbf{EXP3}} \\
DivJac                                & 0.6                               & 0.6                               & 0.65                              & 0.5                               & 0.45                              & 0.55                              & 0.55                              & 0.55                              & 0.45                              & 0.55                              \\
DivBERT                               & 0.93                              & 0.925                             & 0.875                             & 0.875                             & 0.9                               & 0.9                               & 0.95                              & 0.9                               & 0.9                               & 0.875                             \\
DenJac                                & 0.55                              & 0.6                               & 0.55                              & 0.5                               & 0.5                               & 0.6                               & 0.4                               & 0.55                              & 0.6                               & 0.45                              \\
DenBERT                               & 0.92                              & 0.925                             & 0.875                             & 0.875                             & 0.9                               & 0.9                               & 0.9                               & 0.9                               & 0.925                             & 0.9                               \\
TDisJac                               & 0.85                              & 0.85                              & 0.75                              & 0.8                               & 0.75                              & 0.8                               & 0.75                              & 0.8                               & 0.8                               & 0.8                               \\
TDisBERT                              & 0.955                             & 0.955                             & 0.96                              & 0.965                             & 0.96                              & 0.955                             & 0.95                              & 0.955                             & 0.955                             & 0.955                             \\
TDisBLEU                              & 0.55                              & 0.45                              & 0.4                               & 0.35                              & 0.3                               & 0.25                              & 0.15                              & 0.15                              & 0.3                               & 0.25                              \\
TDiff                                 & -1                                & -0.25                             & -1                                & -1.5                              & -1.5                              & -1.5                              & -6                                & -4.5                              & -1.5                              & -1.25                             \\
DivNgram                              & 0.6                               & 0.6                               & 0.55                              & 0.4                               & 0.7                               & 0.7                               & 0.65                              & 0.6                               & 0.75                              & 0.9                               \\
DenNgram                              & 0.00003                           & 0.00004                           & 0.00005                           & 0.00004                           & 0.0001                            & 0.000075                          & 0.00005                           & 0.000075                          & 0.000075                          & 0.00005                          \\
\noalign{\smallskip}\hline
\end{tabular}
\end{table}


\end{landscape}


\appendixsection{Rank Correlation Evolution}
\label{app:rankcorr} 

The following figures depict the evolution of the rank correlation (Kendall's $\tau$) between the shared task's official ranking and the ranking given by the weights learned using online learning. 

\begin{figure}[H]
\centering
\includegraphics[width=1\columnwidth]{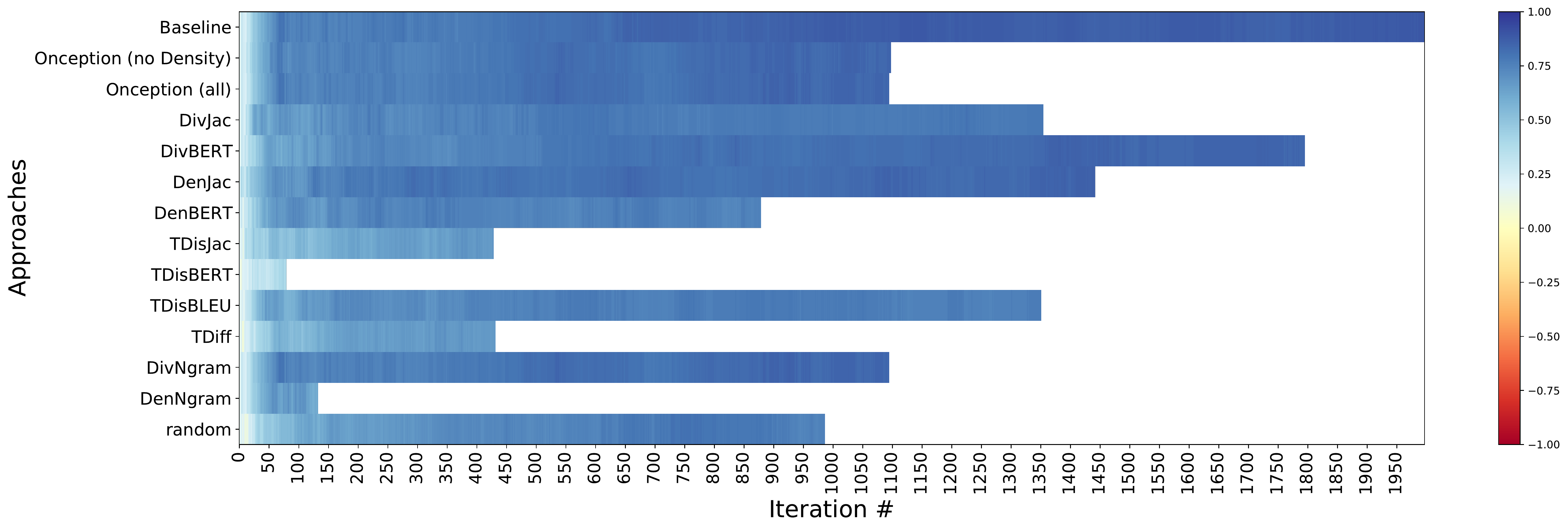}
\caption{Evolution of the rank correlation for {\tt en-de} and \acs{EWAF} (figure best seen in color).}
\label{fig:rankCorrEnDeEWAF}
\end{figure}

\begin{figure}[H]
\centering
\includegraphics[width=1\columnwidth]{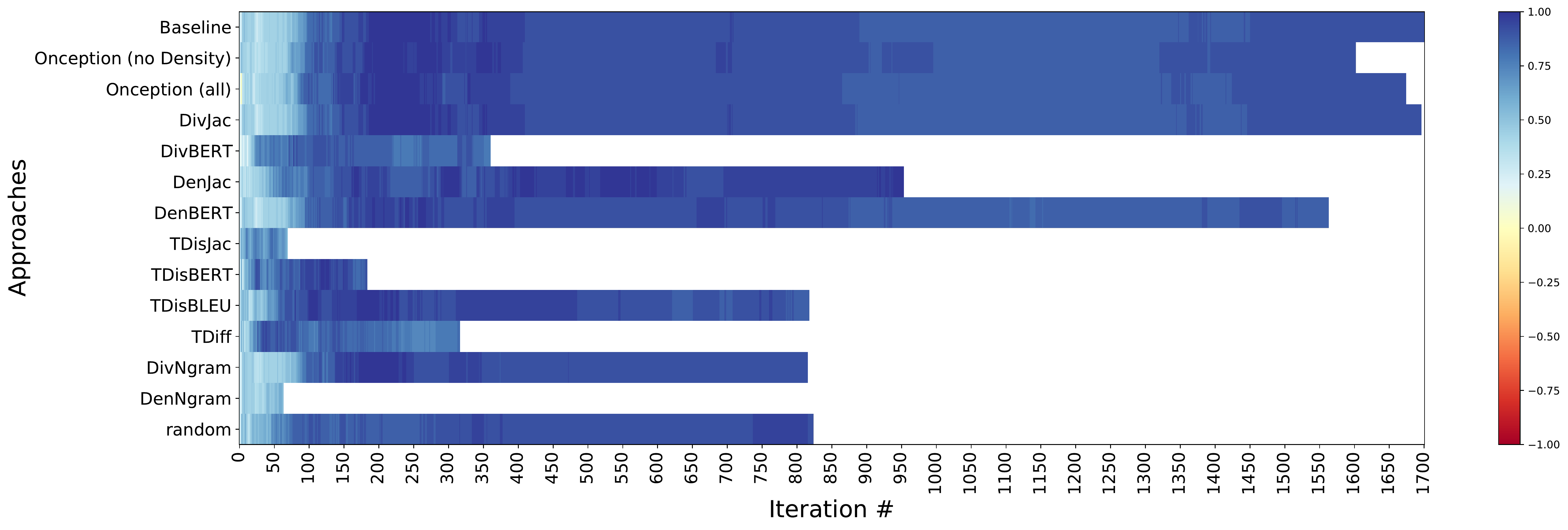}
\caption{Evolution of the rank correlation for {\tt fr-de} and \acs{EWAF} (figure best seen in color).}
\label{fig:rankCorrFrDeEWAF}
\end{figure}

\begin{figure}[H]
\centering
\includegraphics[width=1\columnwidth]{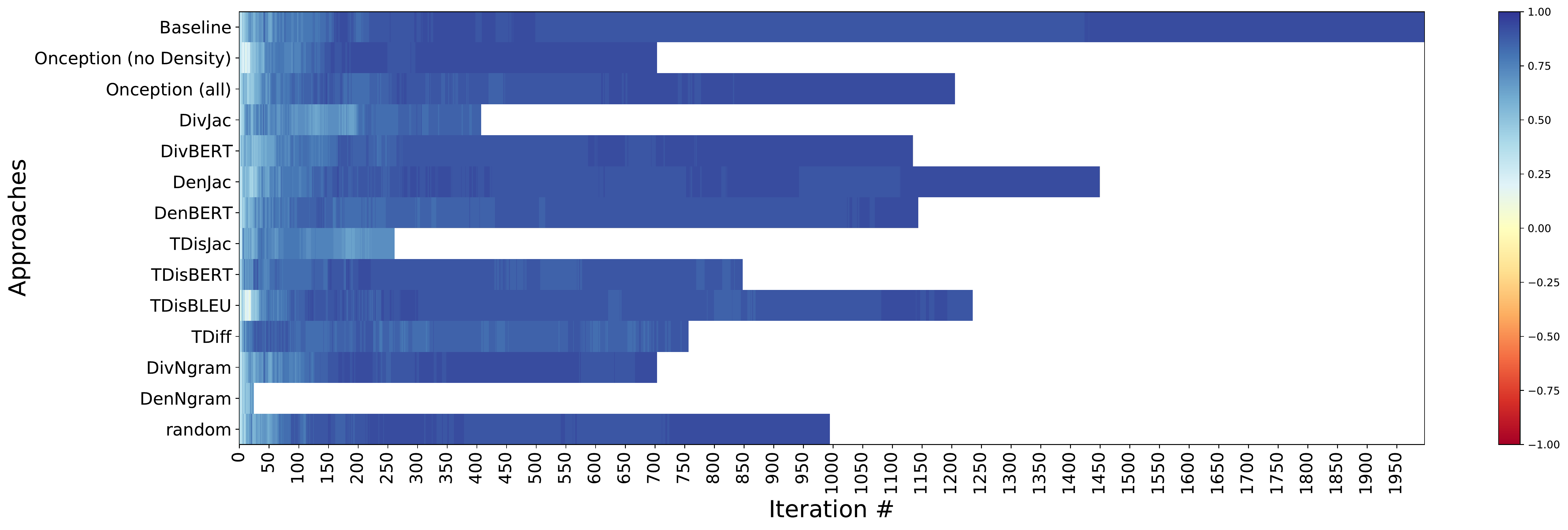}
\caption{Evolution of the rank correlation for {\tt de-cs} and \acs{EWAF} (figure best seen in color).}
\label{fig:rankCorrDeCsEWAF}
\end{figure}

\begin{figure}[H]
\centering
\includegraphics[width=1\columnwidth]{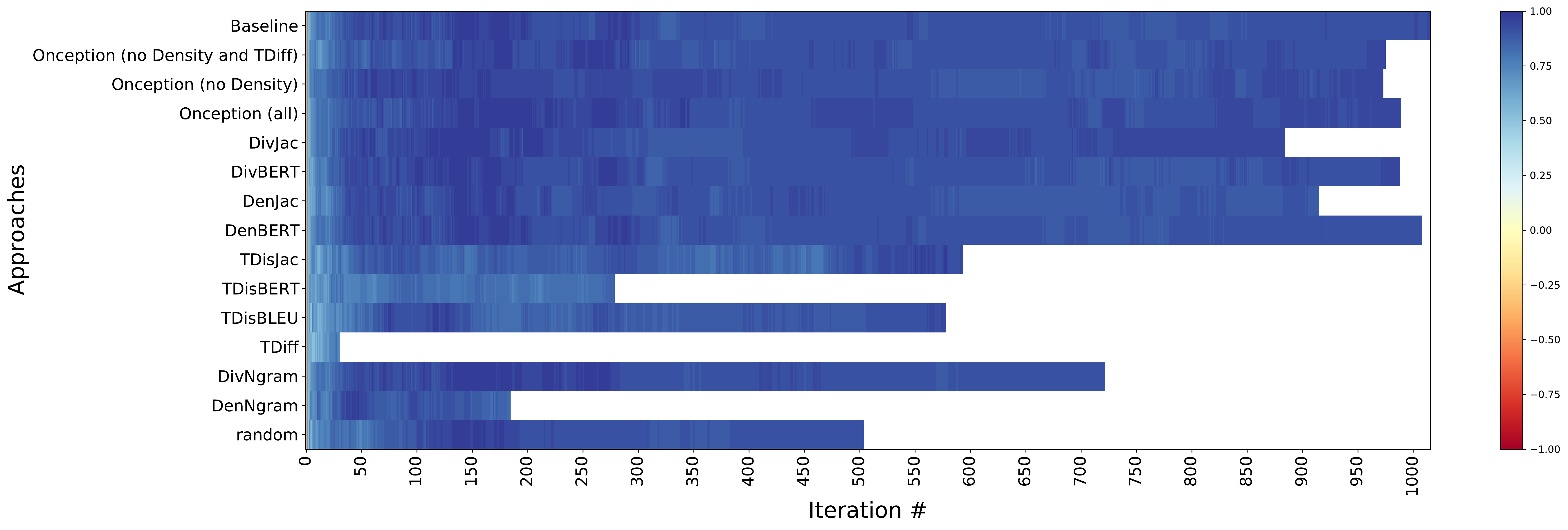}
\caption{Evolution of the rank correlation for {\tt gu-en} and \acs{EWAF} (figure best seen in color).}
\label{fig:rankCorrGuEnEWAF}
\end{figure}

\begin{figure}[H]
\centering
\includegraphics[width=1\columnwidth]{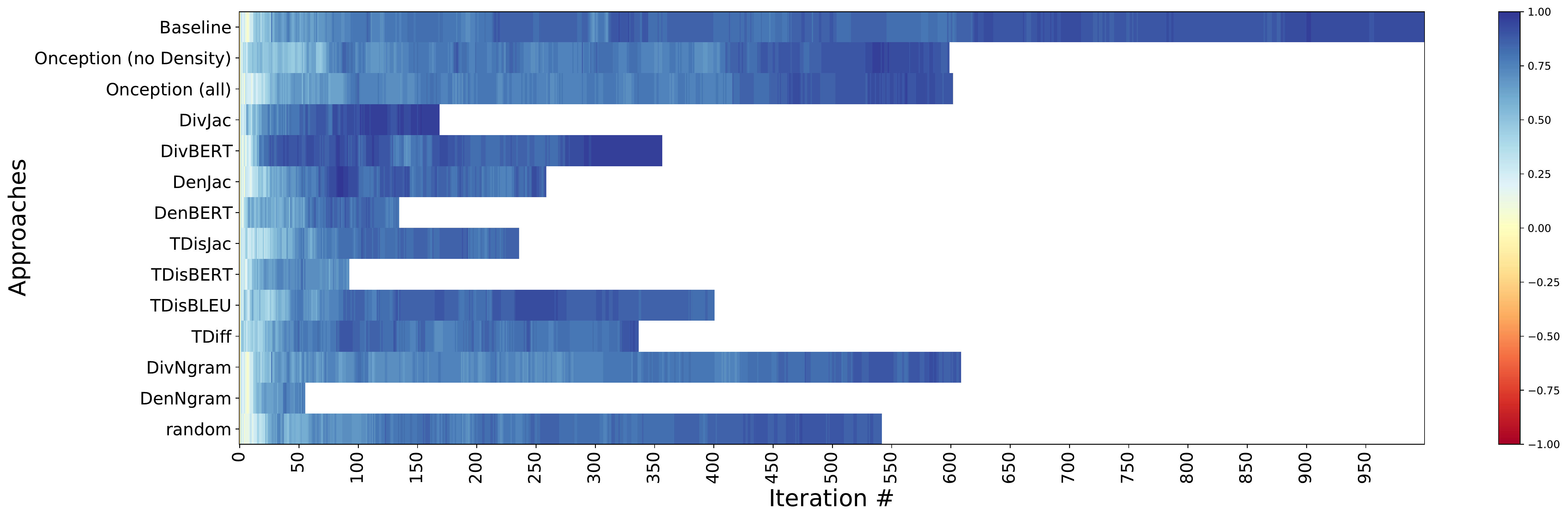}
\caption{Evolution of the rank correlation for {\tt lt-en} and \acs{EWAF} (figure best seen in color).}
\label{fig:rankCorrLtEnEWAF}
\end{figure}

\begin{figure}[H]
\centering
\includegraphics[width=1\columnwidth]{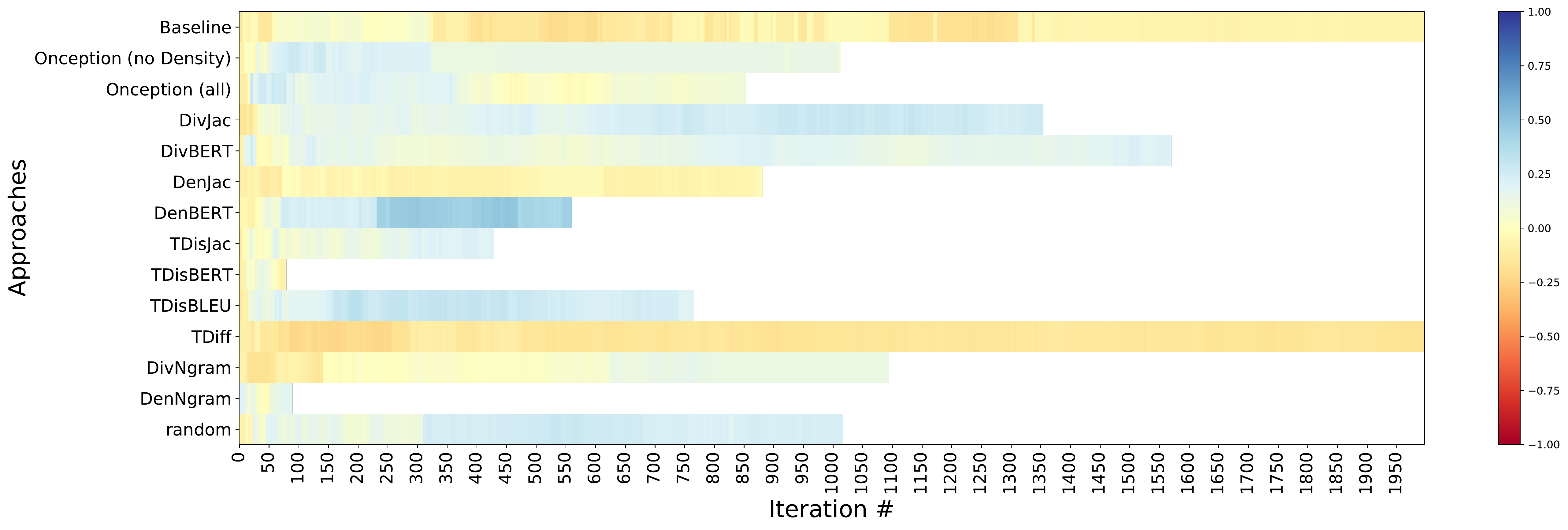}
\caption{Evolution of the rank correlation for {\tt en-de} and \acs{EXP3} (figure best seen in color).}
\label{fig:rankCorrEnDeEXP3}
\end{figure}

\begin{figure}[H]
\centering
\includegraphics[width=1\columnwidth]{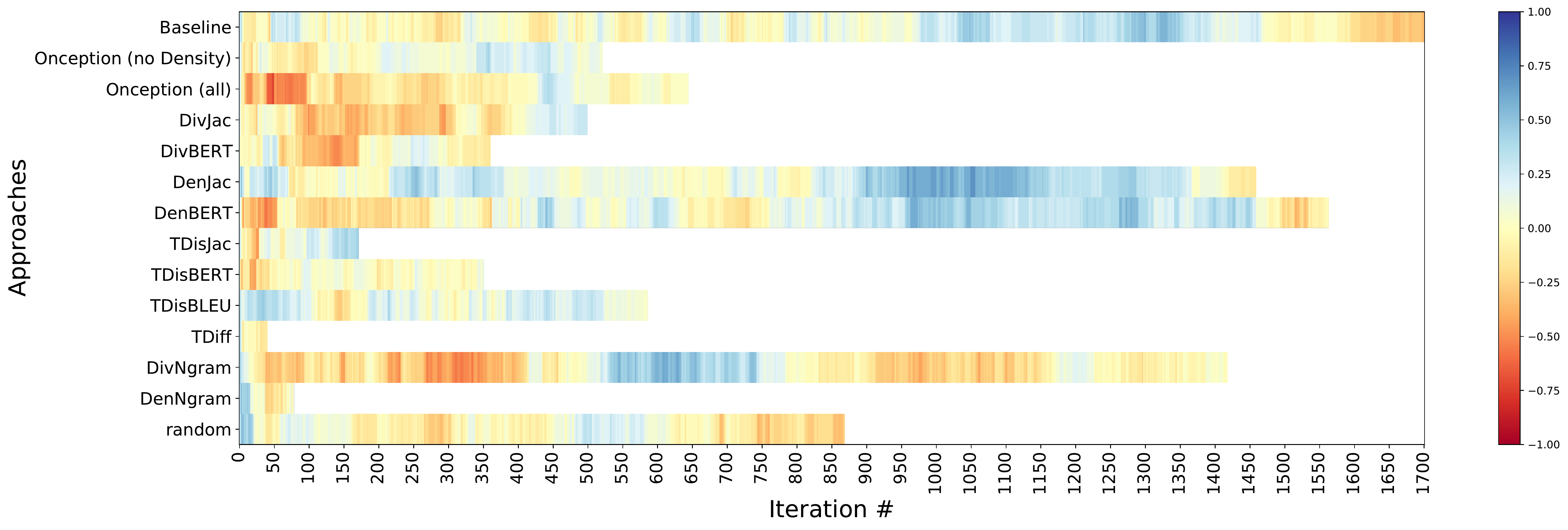}
\caption{Evolution of the rank correlation for {\tt fr-de} and \acs{EXP3} (figure best seen in color).}
\label{fig:rankCorrFrDeEXP3}
\end{figure}

\begin{figure}[H]
\centering
\includegraphics[width=1\columnwidth]{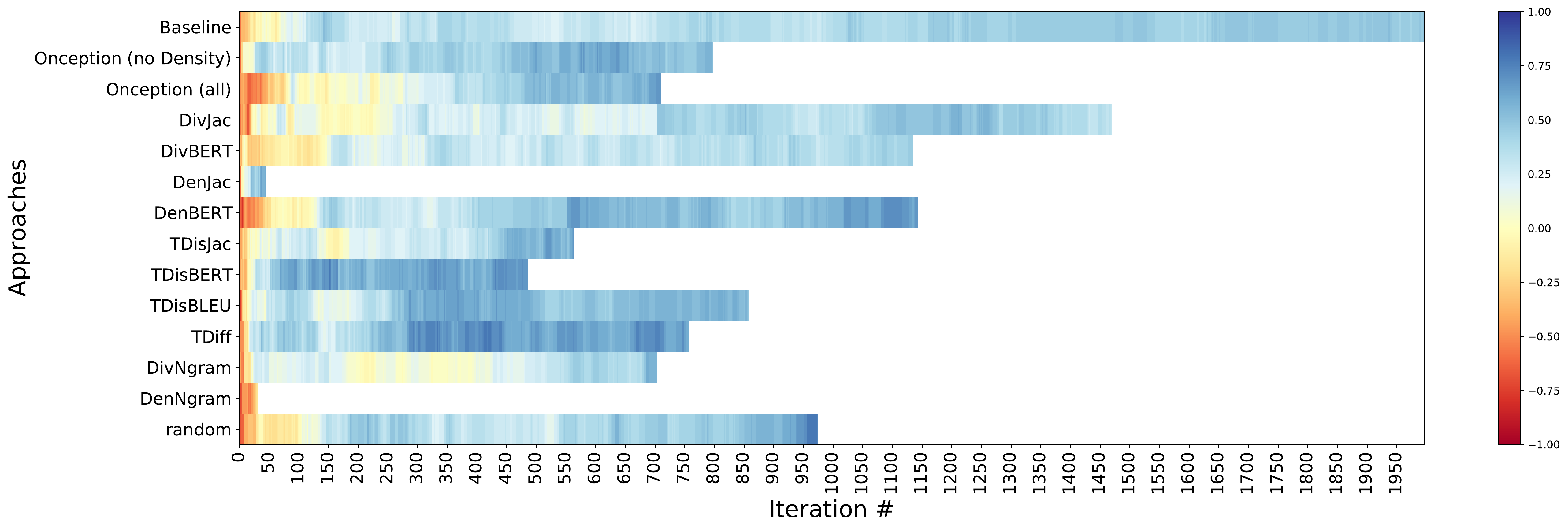}
\caption{Evolution of the rank correlation for {\tt de-cs} and \acs{EXP3} (figure best seen in color).}
\label{fig:rankCorrDeCsEXP3}
\end{figure}

\begin{figure}[H]
\centering
\includegraphics[width=1\columnwidth]{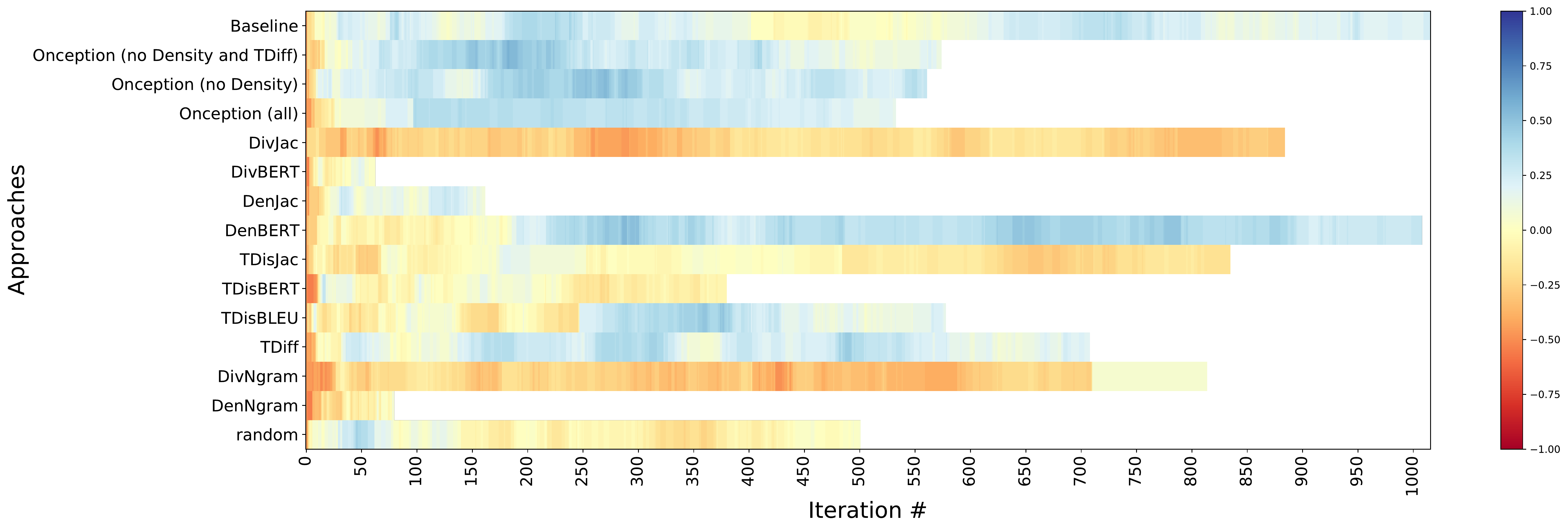}
\caption{Evolution of the rank correlation for {\tt gu-en} and \acs{EXP3} (figure best seen in color).}
\label{fig:rankCorrGuEnEXP3}
\end{figure}

\begin{figure}[H]
\centering
\includegraphics[width=1\columnwidth]{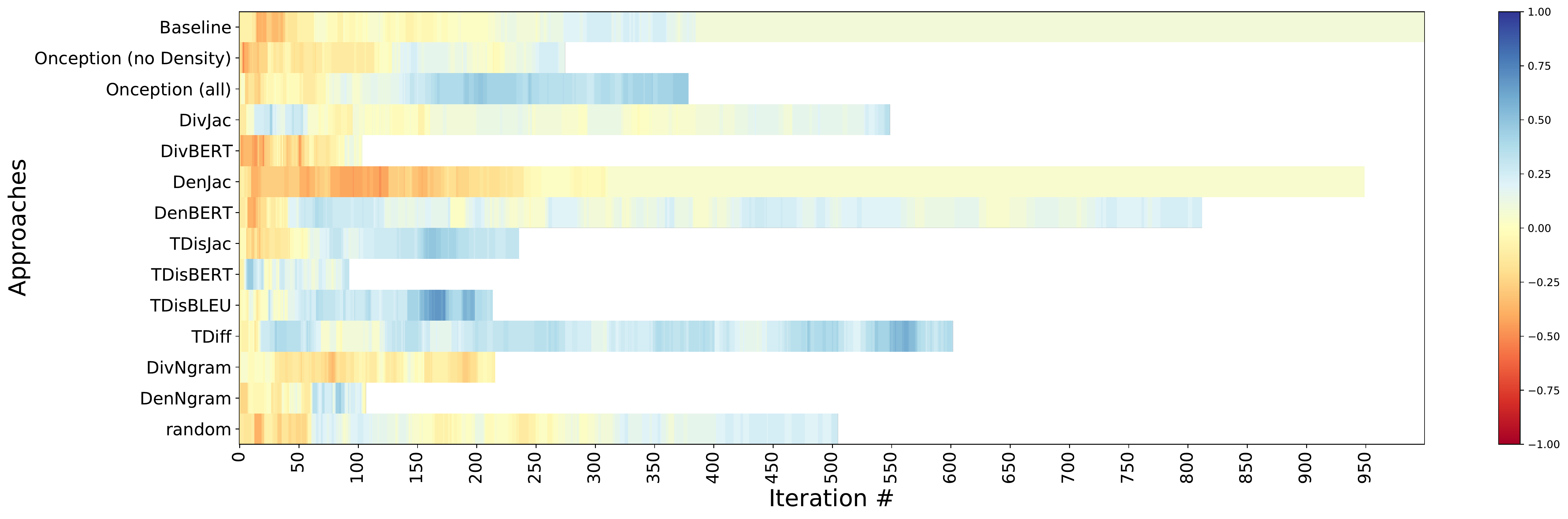}
\caption{Evolution of the rank correlation for {\tt lt-en} and \acs{EXP3} (figure best seen in color).}
\label{fig:rankCorrLtEnEXP3}
\end{figure}

\appendixsection{Query Strategies' Weights Evolution}
\label{app:weightsQS} 

The following figures depict the evolution of the weights of the query strategies when using Onception. 

\begin{figure}[H]
\centering
\includegraphics[width=1\columnwidth]{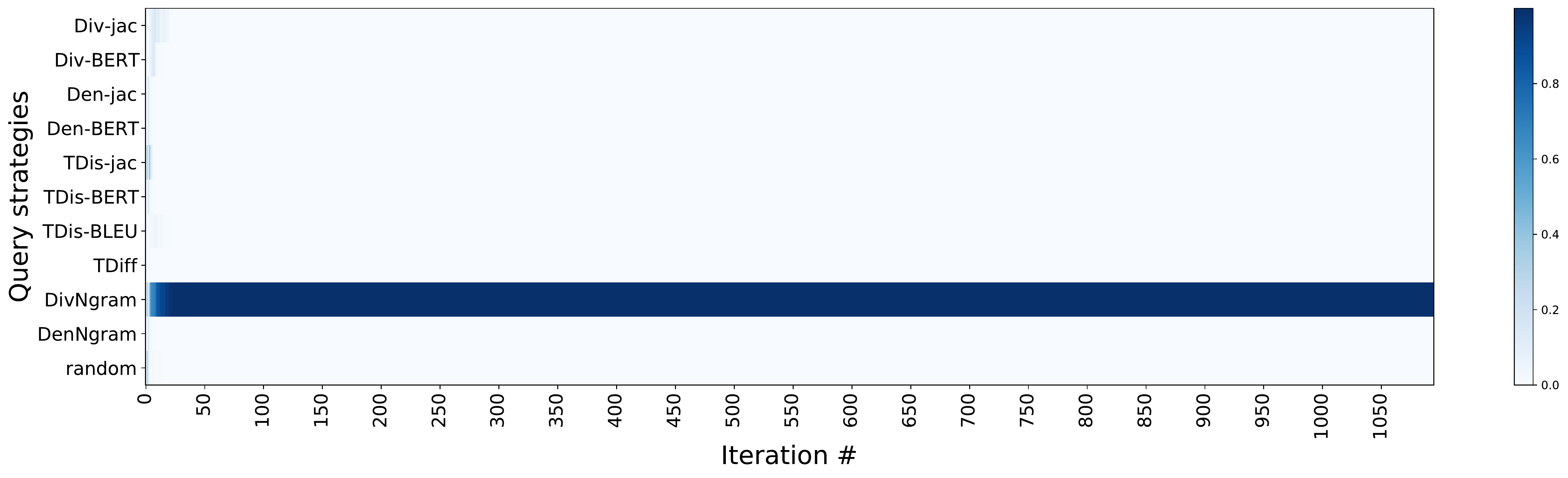}
\caption{Evolution of the query strategies' weights for {\tt en-de} and \acs{EWAF} (figure best seen in color).} 
\label{fig:weightsQSEnDeEWAF}
\end{figure}

\begin{figure}[H]
\centering
\includegraphics[width=1\columnwidth]{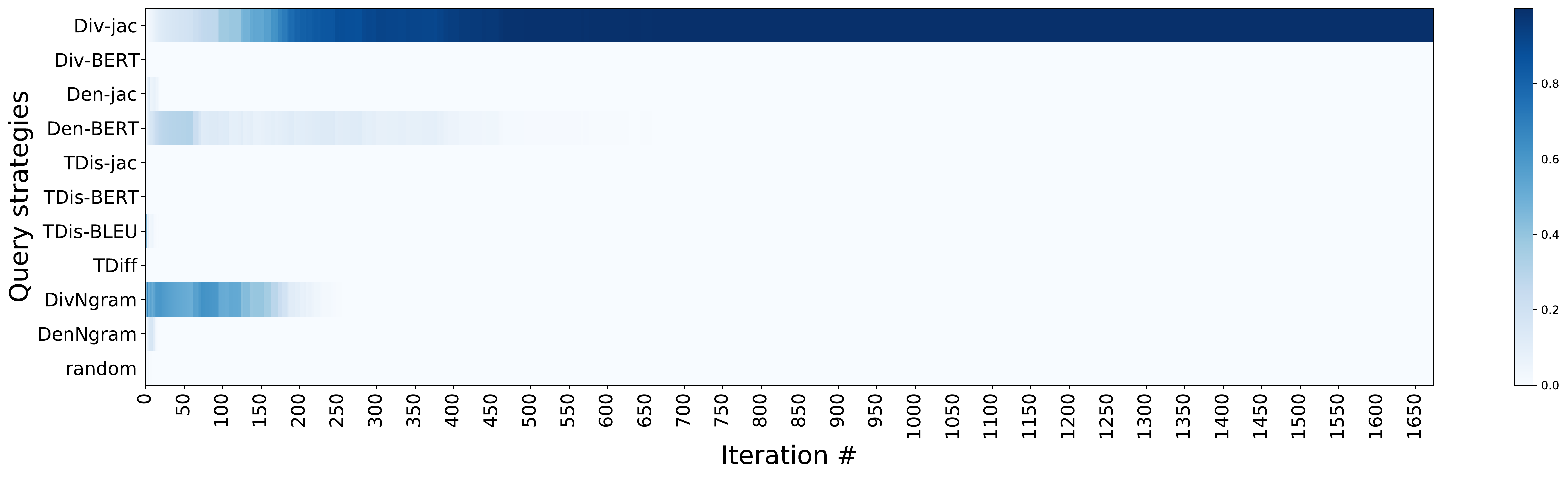}
\caption{Evolution of the query strategies' weights for {\tt fr-de} and \acs{EWAF} (figure best seen in color).}
\label{fig:weightsQSFrDeEWAF}
\end{figure}

\begin{figure}[H]
\centering
\includegraphics[width=1\columnwidth]{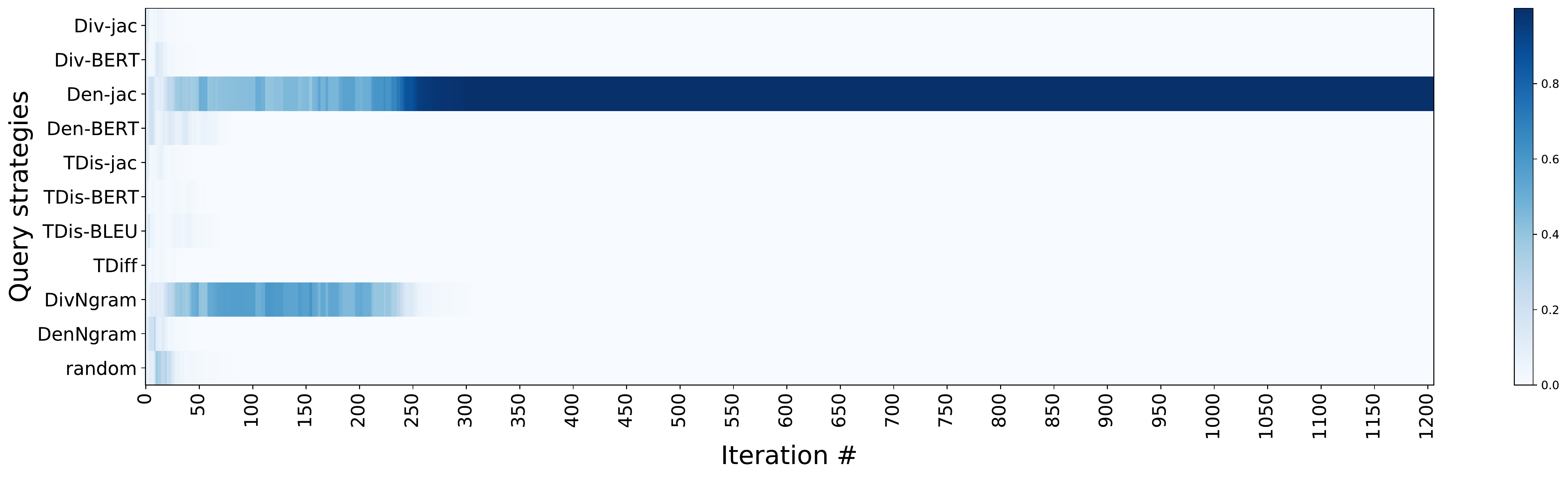}
\caption{Evolution of the query strategies' weights for {\tt de-cs} and \acs{EWAF} (figure best seen in color).}
\label{fig:weightsQSDeCsEWAF}
\end{figure}

\begin{figure}[H]
\centering
\includegraphics[width=1\columnwidth]{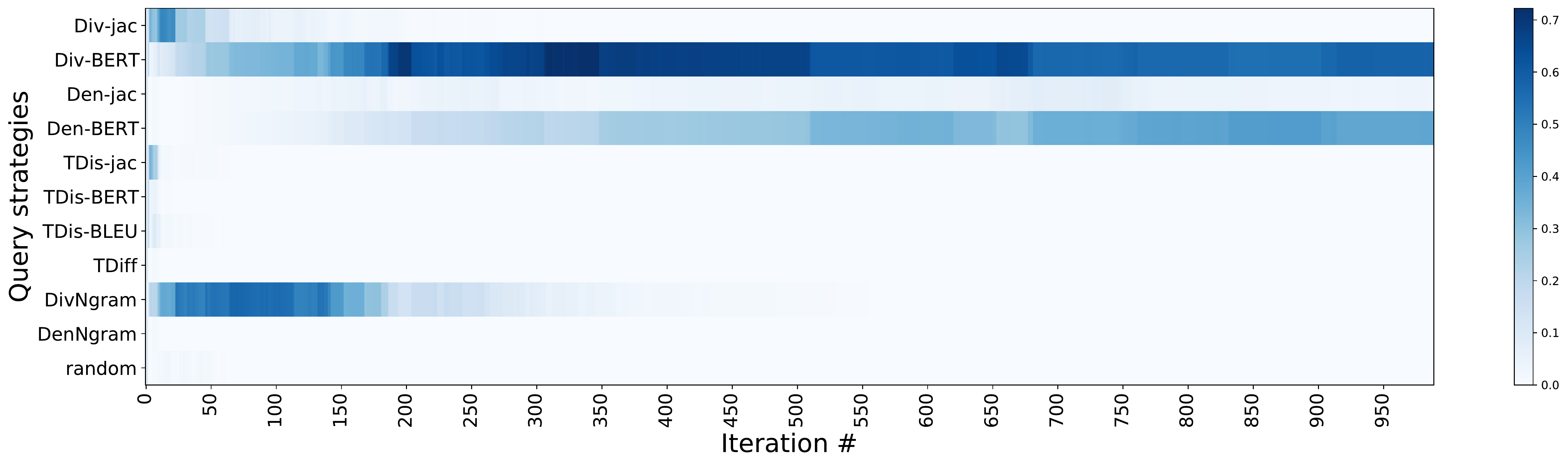}
\caption{Evolution of the query strategies' weights for {\tt gu-en} and \acs{EWAF} (figure best seen in color).}
\label{fig:weightsQSGuEnEWAF}
\end{figure}

\begin{figure}[H]
\centering
\includegraphics[width=1\columnwidth]{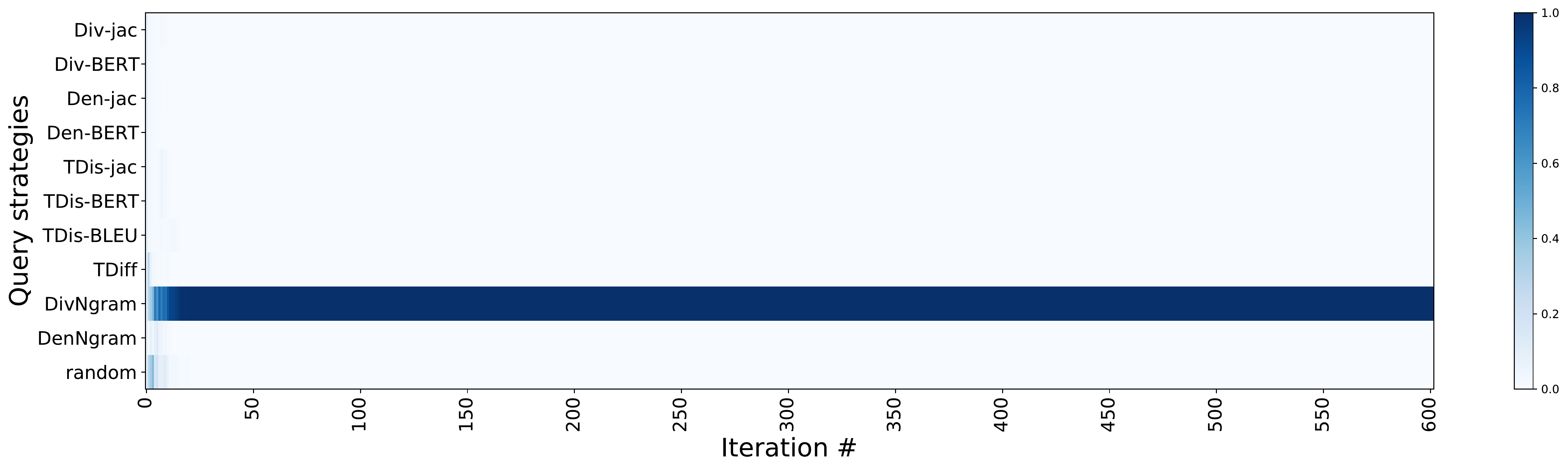}
\caption{Evolution of the query strategies' weights for {\tt lt-en} and \acs{EWAF} (figure best seen in color).}
\label{fig:weightsQSLtEnEWAF}
\end{figure}

\begin{figure}[H]
\centering
\includegraphics[width=1\columnwidth]{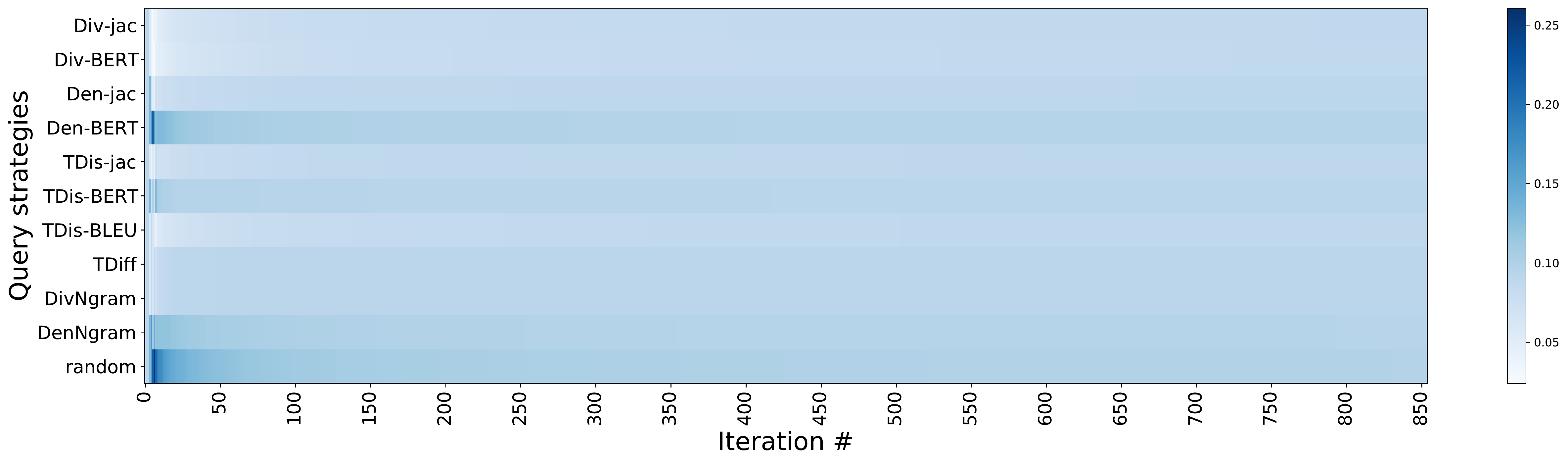}
\caption{Evolution of the query strategies' weights for {\tt en-de} and \acs{EXP3} (figure best seen in color).}
\label{fig:weightsQSEnDeEXP3}
\end{figure}

\begin{figure}[H]
\centering
\includegraphics[width=1\columnwidth]{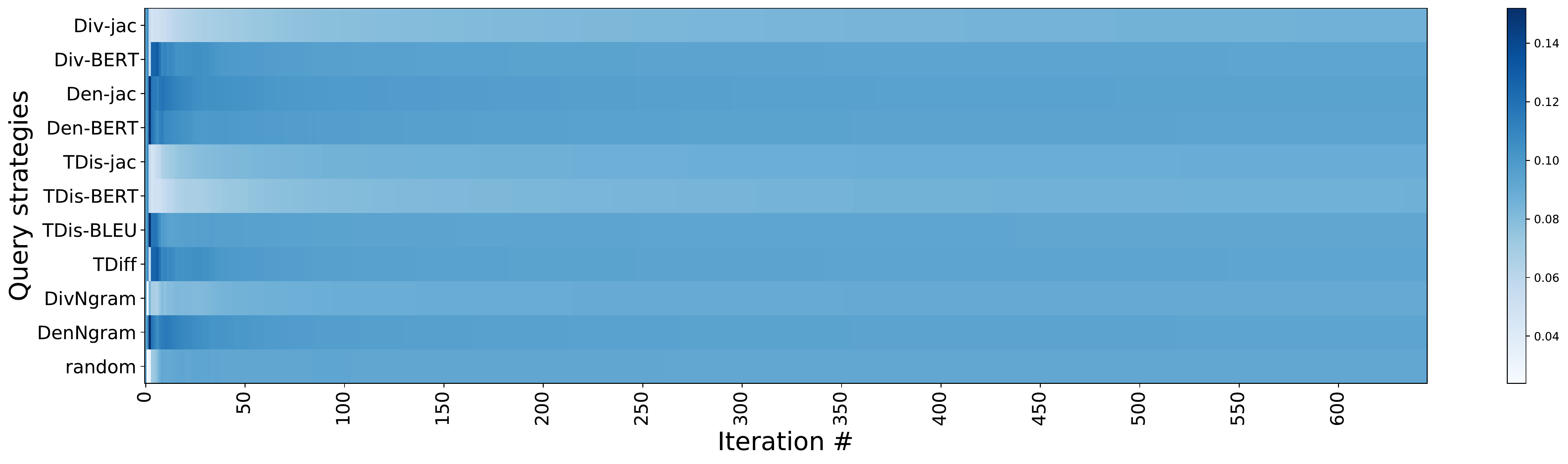}
\caption{Evolution of the query strategies' weights for {\tt fr-de} and \acs{EXP3} (figure best seen in color).}
\label{fig:weightsQSFrDeEXP3}
\end{figure}

\begin{figure}[H]
\centering
\includegraphics[width=1\columnwidth]{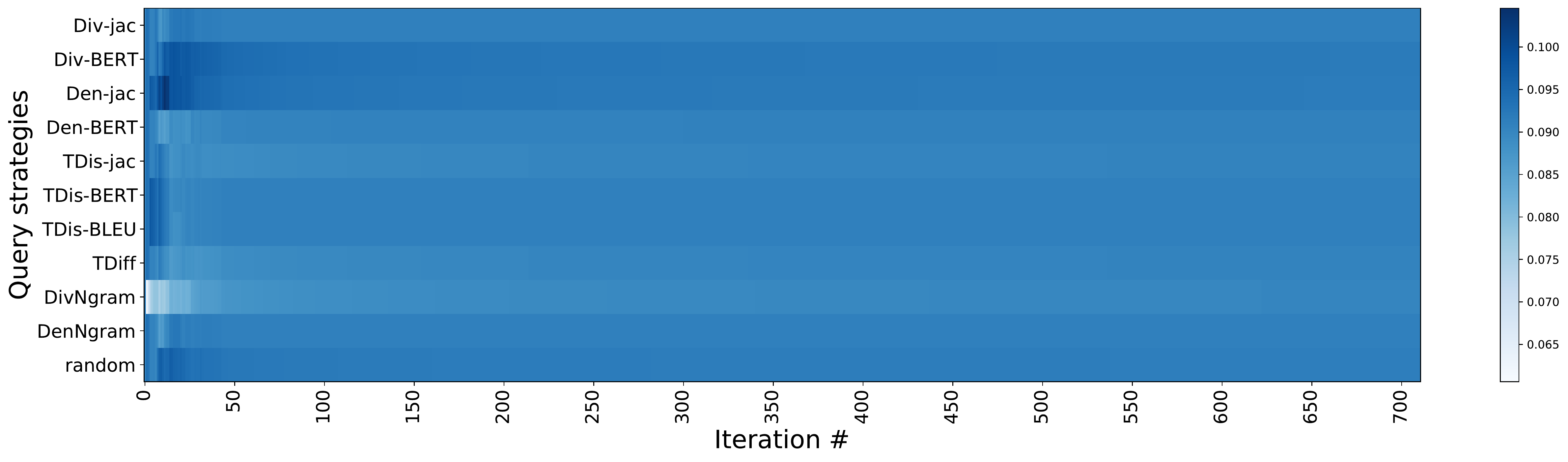}
\caption{Evolution of the query strategies' weights for {\tt de-cs} and \acs{EXP3} (figure best seen in color).}
\label{fig:weightsQSDeCsEXP3}
\end{figure}

\begin{figure}[H]
\centering
\includegraphics[width=1\columnwidth]{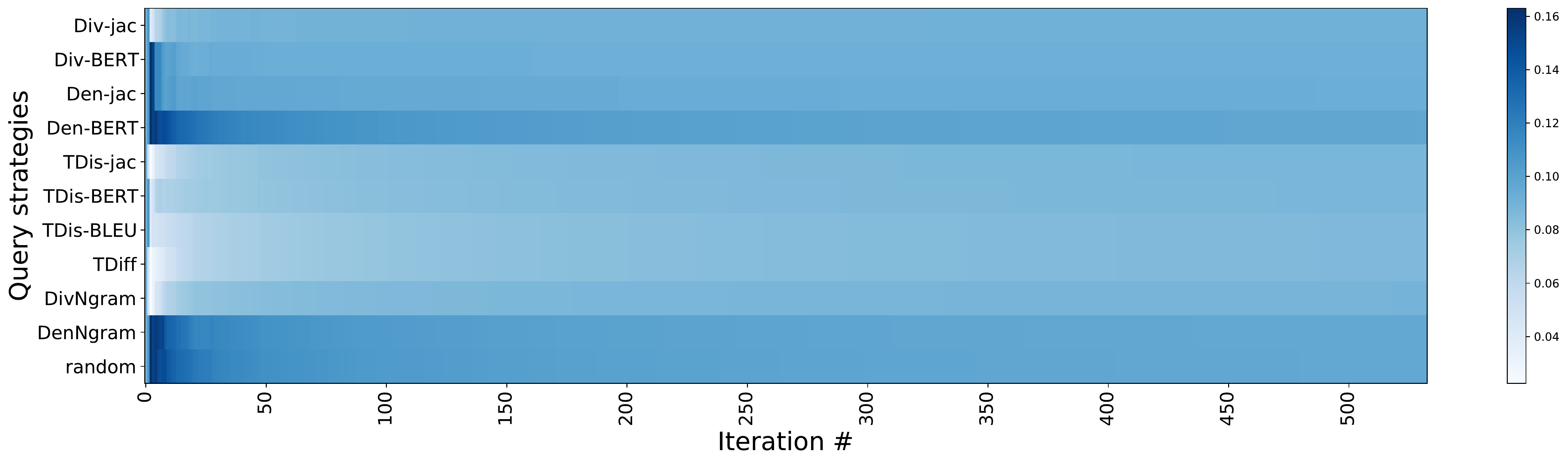}
\caption{Evolution of the query strategies' weights for {\tt gu-en} and \acs{EXP3} (figure best seen in color).}
\label{fig:weightsQSGuEnEXP3}
\end{figure}

\begin{figure}[H]
\centering
\includegraphics[width=1\columnwidth]{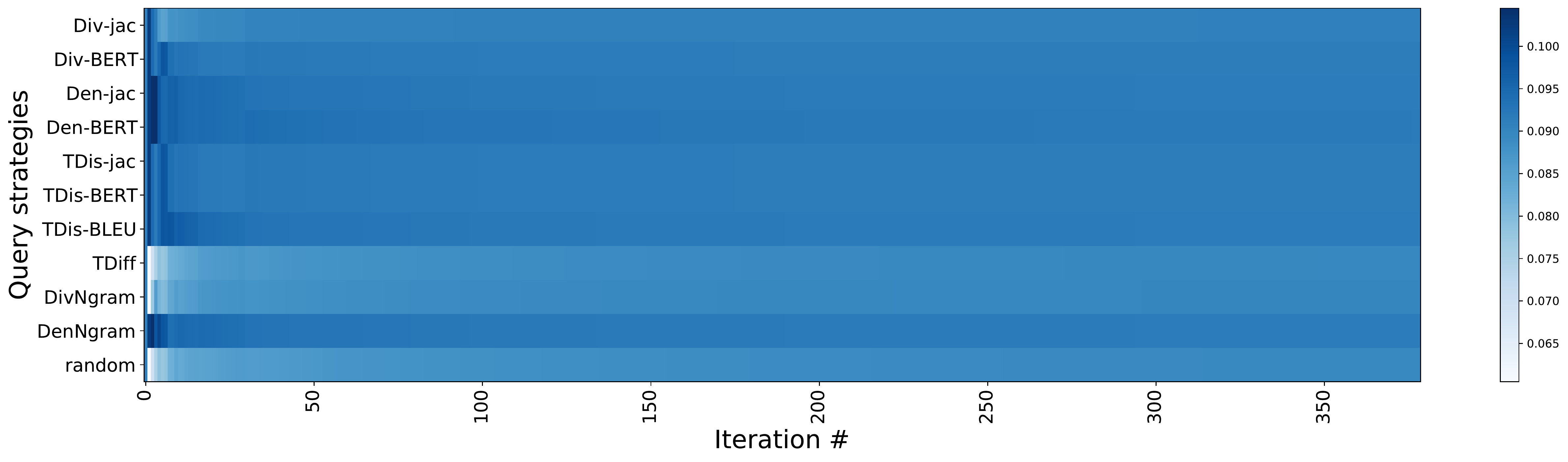}
\caption{Evolution of the query strategies' weights for {\tt lt-en} and \acs{EXP3} (figure best seen in color).}
\label{fig:weightsQSLtEnEXP3}
\end{figure}

\newpage

\begin{acknowledgments}
This work was supported by: Fundação para a Ciência e a Tecnologia under reference UIDB/50021/2020 (INESC-ID multi-annual funding), as well as under the HOTSPOT project with reference PTDC/CCI-COM/7203/2020; P2020 program, supervised by Agência Nacional de Inovação (ANI), under the project CMU-PT Ref. 045909 (MAIA). 

Vânia Mendonça was funded by Fundação para a Ciência e a Tecnologia with a PhD grant, ref. SFRH/BD/121443/2016.

The authors would like to thank to Ana Lúcia Santos for proof-reading the document, and Soraia M. Alarcão for helping during brainstorming. 
\end{acknowledgments}


\starttwocolumn
\bibliography{CL-Onception}


\end{document}